\documentclass[11pt]{article}
\usepackage[margin=1in]{geometry}

\usepackage{cite}
\usepackage{amsmath,amssymb,amsfonts}

\usepackage{graphicx}
\usepackage{booktabs}
\usepackage[linesnumbered,ruled,vlined]{algorithm2e}
\usepackage{hyperref}
\usepackage{orcidlink}
\hypersetup{hidelinks=true}
\usepackage{textcomp}
\usepackage{amsthm}

\title{Enhancing Federated Survival Analysis through Peer-Driven Client Reputation in Healthcare}

\author{
Navid Seidi\,\orcidlink{0000-0002-9912-0103}\thanks{Department of Computer Science, Missouri University of Science and Technology, Rolla, MO, USA. Email: \texttt{nseidi@mst.edu}} \and
Satyaki Roy\,\orcidlink{0000-0001-6767-266X}\thanks{Department of Mathematical Sciences, University of Alabama in Huntsville, Huntsville, AL, USA. Email: \texttt{sr0215@uah.edu}} \and
Sajal Das\,\orcidlink{0000-0002-9471-0868}\thanks{Department of Computer Science, Missouri University of Science and Technology, Rolla, MO, USA. Email: \texttt{sdas@mst.edu}}
}

\date{}  

\begin{document}

\maketitle

\begin{abstract}
Federated Learning (FL) holds great promise for digital health by enabling collaborative model training without compromising patient data privacy. However, heterogeneity across institutions, lack of sustained reputation, and unreliable contributions remain major challenges. In this paper, we propose a robust, peer-driven reputation mechanism for federated healthcare that employs a hybrid communication model to integrate decentralized peer feedback with clustering-based noise handling to enhance model aggregation. Crucially, our approach decouples the federated aggregation and reputation mechanisms by applying differential privacy to client-side model updates before sharing them for peer evaluation. This ensures sensitive information remains protected during reputation computation, while unaltered updates are sent to the server for global model training. Using the Cox Proportional Hazards model for survival analysis across multiple federated nodes, our framework addresses both data heterogeneity and reputation deficit by dynamically adjusting trust scores based on local performance improvements measured via the concordance index. Experimental evaluations on both synthetic datasets and the SEER dataset demonstrate that our method consistently achieves high and stable C-index values, effectively down-weighing noisy client updates and outperforming FL methods that lack a reputation system.
\end{abstract}


\section{Introduction}
\label{sec:intro}
Federated Learning (FL) has emerged as a transformative paradigm for distributed model training, enabling multiple clients to collaboratively learn a shared model without the need to exchange raw data \cite{mcmahan2017communication}. FL has particular appeal in sensitive domains such as healthcare, where privacy concerns and regulatory requirements prohibit centralized data pooling. Despite its promise, FL confronts significant challenges, notably in terms of a lack of established reputation, data heterogeneity, and unreliable contributions. In real-world healthcare environments, institutions—such as hospitals and clinics— often engage in both collaborative and competitive interactions~\cite{goddard2015competition}. However, institute-specific data characteristics have been shown to significantly influence analytical outcomes~\cite{seidi2023using}. As such, ensuring a robust and fair FL process necessitates mechanisms that assess and update the evolving reliability.

Recent advances have demonstrated that clustering clients based on similar disease profiles can improve survival prediction in heterogeneous healthcare datasets \cite{ouyang2022clusterfl,kim2021dynamic,seidi2024addressing}. However, there remains a need to more effectively leverage such clustering strategies, particularly by forming homogeneous groups aligned with feature completeness and risk profiles. Properly addressing this gap is essential for improving the reliability and performance of aggregated models in healthcare FL.

The \textit{second} major challenge in federated model aggregation is designing reliable mechanisms for assessing client contributions without relying on centralized authorities. Existing methods often depend on a central mechanism and a trusted dataset to compute reputation scores \cite{kang2019incentive, wang2020novel, rashid2025trustworthy}, which introduces vulnerabilities and scalability concerns. While feedback-based peer evaluation offers a promising alternative, effective decentralized reputation systems that dynamically adjust based on observed improvements in local model performance remain underdeveloped. Specifically, \textit{reputation} is defined as the degree of confidence a client has in the reliability and quality of a peer’s contributions, quantifiable through performance metrics comparing accuracy with and without a peer’s updates. 

\textit{Third}, data sensitivity concerns also present a major hurdle for peer-driven model aggregation in FL, particularly in sensitive domains like healthcare. Sharing model parameters directly with neighboring clients risks leaking sensitive information about local datasets. Although \textit{differential privacy} techniques have been developed to mitigate such risks, integrating robust privacy mechanisms without compromising the utility of peer evaluation remains an open challenge~\cite{abadi2016deep,dwork2006calibrating}. The \textit{fourth} concern involves modeling the evolution of client behavior over time, particularly under adversarial conditions. Most existing works assume static client behavior; however, in realistic federated settings, clients may initially behave honestly to establish trust before gradually introducing malicious updates. In this regard, dynamic threat modeling with parameters to modulate the degree of adversarial noise injected into feature vectors has remained underexplored. 

In keeping with prior work on the importance of trust and incentive mechanisms in collaborative networks~\cite{kang2019incentive, wang2020novel, rashid2025trustworthy}, our framework advances reputation-aware federated learning for healthcare applications. A key innovation lies in the use of \textit{differential privacy} to decouple the peer-based reputation system from the core federated aggregation process: client updates are privatized before being shared with peers for evaluation, while unaltered versions are submitted to the server for global model training. This separation not only safeguards sensitive model parameters during reputation computation but also allows for independent and secure peer interactions without compromising convergence or model quality. Additionally, the method unifies strategies for managing data heterogeneity and assessing client reliability by employing a dynamic reputation score that reflects local model improvements based on the concordance index. Our threat model considers stealthy adversaries that gradually degrade update quality, and we incorporate noise-resilient peer feedback through clustering. Experiments on both synthetic and SEER datasets confirm that our approach delivers robust models, outperforming standard FL methods, namely, Federated Averaging~\cite{mcmahan2017communication} and Trustworthy and Fair Federated Learning~\cite{rashid2025trustworthy}, lacking trust-awareness.

\section{Background and Related Work}
\label{sec:background}

\subsection{Survival Analysis}\label{sec:coxph}
Survival analysis is a statistical technique for analyzing the time until an event of interest. e.g., death, healing, disease recurrence, etc. A standard approach for survival analysis is the \textit{Cox proportional hazards} (CoxPH) model~\cite{cox1972regression} that estimates the hazard rate, which is the instantaneous rate of occurrence of an event at a point in time, given that no event has occurred till that time, contingent upon various covariates \(x\). 
The hazard function of the CoxPH model for an individual is defined by:
\begin{equation}
\label{eq:CoxPH_hazard}
    h(t|x) = h_0(t) \exp(\beta^T x),
\end{equation}
Here, \(h_0(t)\) denotes the baseline hazard function describing the hazard rate, serving as a benchmark against which the effects of covariates are measured. The vector \(\beta\) quantifies the influence of a covariate on the hazard rate. These coefficients are a measure of the relative risk associated with the covariates. In our federated framework, we represent the locally estimated survival model coefficients \(\beta\) collectively as \(\theta\), which are shared with and aggregated by the central server.

\subsection{Central Server-Based Federated Learning Architecture}
Conventional Federated Learning (FL) in healthcare predominantly employs a centralized architecture, in which a single server coordinates the training process across distributed client nodes. In this model, the server initializes a global model and disseminates it to the clients; each client then trains the model on its private data and sends back updated parameters or gradients. The server aggregates these updates—typically using Federated Averaging (FedAvg) \cite{mcmahan2017communication} - so that nodes with larger datasets exert a proportionately greater influence on the global model. This centralized scheme simplifies synchronization and allows for effective monitoring of training progress and anomaly detection. However, it also introduces critical vulnerabilities: a central server represents a single point of failure and is susceptible to hardware malfunctions, cyber-attacks, or connectivity disruptions \cite{chen2024trustworthy}. Furthermore, as the number of clients grows, the server becomes a communication and computation bottleneck, which limits overall system scalability. Reputation, trust, and privacy concerns are also prominent because the central server must be fully trusted not only to correctly aggregate model updates but also to safeguard sensitive domain-specific, namely, healthcare, data. The use of privacy-preserving techniques, such as secure aggregation or differential privacy, adds computational overheads~\cite{fang2024byzantine, zhou2024defta}.

\subsection{Decentralized Reputation Mechanisms}
\label{sec:dec}

The studies led by Blanchard et al. and Yin et al. address the resilience of distributed machine learning algorithms to Byzantine failures~\cite{blanchard2017machine,yin2018byzantine}. Blanchard et al. introduce techniques to mitigate the impact of arbitrary failures in distributed stochastic gradient descent (SGD) implementations, enhancing the robustness of machine learning systems against adversarial behaviors. Similarly, Yin et al. develop distributed optimization algorithms that are provably robust against Byzantine failures, aiming to achieve optimal statistical performance despite adversarial nodes. El Mhamdi et al. show that the critical vulnerability in existing Byzantine-resilient SGD methods is that the adversaries can exploit high-dimensional loss landscapes to steer model convergence toward ineffective solutions. They propose a robust aggregation rule that significantly reduces the adversarial influence, achieving near-optimal model updates while maintaining overall convergence integrity~\cite{guerraoui2018hidden}.

\subsection{Concordance Index}
\label{sec:CI}
The \emph{concordance index} (C-index)\cite{harrell2001regression} is a widely used metric for assessing the discriminatory ability of survival models. It quantifies the probability that, for a randomly chosen pair of subjects, the individual with a higher predicted risk, or equivalently, a lower predicted survival time) experiences the event before the other. Formally, if $f(x)$ denotes the risk score for a subject with covariates $x$, the C-index is defined as:
\begin{equation}
    C = \Pr\Bigl( f(x_i) > f(x_j) \,\big|\, t_i < t_j \Bigr),
\end{equation}
\noindent In the above equation, $t_i$ and $t_j$ denote the observed event times of subjects $i$ and $j$, respectively. The C-index provides an intuitive measure of a model's ability to correctly rank subjects by risk, with a value of \( 0.5 \) indicating no better than random chance and a value of \( 1 \) denoting perfect discrimination.

\section{Reputation Mechanism Framework}\label{sec:repute}

We define the \textit{reputation} of client \( i \) with respect to peer \( j \), denoted \( RS_{ij}(t) \), as a dynamic measure of peer-assessed confidence in the quality of \( i \)’s model updates. It is updated based on improvements in predictive performance (e.g., concordance index) attributed to \( i \)’s contributions, supporting robust aggregation in heterogeneous, evolving settings.

\subsection{Communication Model}
In a hybrid communication scenario, model aggregation and reputation updates follow bidirectional communication paradigms. Model aggregation adheres to the FL framework, where clients exchange model updates directly with the server in a client-server architecture. Parallelly, reputation updates occur through peer-to-peer (P2P) communication, where two clients share assessments of a third client, allowing reputation values to evolve dynamically based on collective evaluations. Overall, this model balances global coordination with local adaptability, strengthening the integrity in adversarial settings.

\subsection{Problem Formulation}\label{sec:form}
The objective function involves optimizing the assignment of data to clusters while assigning a data point to the cluster that minimizes the Euclidean distance between the point and the cluster centroid while also maximizing the concordance (see Section \ref{sec:CI}) between predicted risk scores and observed survival times. This \textit{clustering-concordance trade-off} is achieved through the regularization parameter \( \lambda \), which controls the importance of the intra-cluster concordance index.

\[
\min_{C_1, C_2, \dots, C_c} \{ \sum_{i=1}^{c} \sum_{j \in C_i} \| B_j - \mu_i \|_2 \;\]
\[- \; \lambda \cdot \sum_{i=1}^{c} \sum_{(j, \ell) \in P_i} \mathbb{I}\Bigl( (r_j - r_\ell)(t_\ell - t_j) > 0 \Bigr) \}
\]

The terms in the objective function are explained as follows. \( C_1, C_2, \dots, C_c \) represent the clusters, with \( c \) denoting the total number of clusters. The objective is to optimize the assignment of data points to these clusters. \( B_j \) refers to the \( j \)-th data point in the bulk matrix \( B \); here, \( B_j \) is a positive, real-valued feature vector where each entry corresponds to the proportion of non-null values (i.e. the completeness) for a global feature. \( \mu_i \) denotes the centroid of cluster \( C_i \), computed as the mean of all feature vectors of data points assigned to \( C_i \). The term \( \| B_j - \mu_i \|_2 \) is the Euclidean distance between the data point \( B_j \) and the cluster centroid \( \mu_i \), which encourages grouping centers with similar patterns of feature completeness. As stated earlier, the parameter \( \lambda \) controls the importance of the C-index.

The set \( P_i \) consists of the permissible pairs of patients within cluster \( i \); a pair \( (j, \ell) \) is considered permissible if there is a valid comparison between patients \( j \) and \( \ell \) based on their observed survival times. The predicted risk scores \( r_j \) and \( r_\ell \) correspond to patients \( j \) and \( \ell \), respectively - with higher values of \( r_j \) indicating a higher risk of an event (e.g., death) for patient \( j \). Similarly, \( t_j \) and \( t_\ell \) represent the observed survival times for patients \( j \) and \( \ell \), where lower values of \( t_j \) indicate shorter survival times for patient \( j \). Lastly, the indicator function, \( \mathbb{I}\Bigl( (r_j - r_\ell)(t_\ell - t_j) > 0 \Bigr) \), returns a value of \( 1 \) if the ordering of predicted risk scores matches the ordering of survival times (i.e., the pair is concordant), and 0 otherwise.

\subsection{Threat Model}\label{sec:threat}
The threat model reflects a scenario where a client node may initially behave honestly, before progressively adding noise to its feature vector after a predefined time, \( T^{honest} \). The evolution of the feature vector \( v_t \) is governed by:
\[
v_t = 
\begin{cases} 
v_t^{\text{honest}}, & \text{if } t < T^{\text{honest}} \\
v_t^{\text{honest}} + n_t, & \text{otherwise.}
\end{cases}
\]

\noindent Here, \( n_t \) is the noise added to the feature vector, defined as:
\( n_t = \alpha_i \cdot z_i, \quad \text{where } \alpha_i = \min\left(\frac{t - T^{\text{honest}}}{T^{\text{ramp}}}, \epsilon_{\max}\right) \). Note that the parameter \( \alpha_i \) controls the scaling of the noise, with a ramp-up period defined by \( T^{\text{ramp}} \), during which the noise grows gradually until it reaches a maximum allowable magnitude, \( \epsilon_{\max} \). The vector \( z_i \) represents the noise direction, chosen such that \( \sum_i z_i = 0 \). A lower value of \( T^{\text{honest}} \) results in an abrupt injection of noise shortly after training begins, whereas a higher \( T^{\text{honest}} \) leads to a gradual introduction of noise.

\begin{figure}[h!]
    \centering    
    \includegraphics[width=0.5\textwidth]{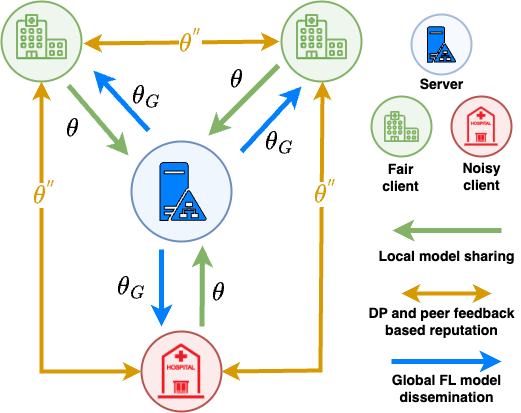}
    \caption{The proposed approach combining server-client and peer-to-peer data flow for model aggregation and reputation updates.}
    \label{fig:schematic}
\end{figure}

\subsection{Message Exchange}
For client $i$, the feedback from client $j$ regarding client $k$'s contribution is defined as: \( m_{j,k} = \Omega(\mathcal{M}_k) - \Omega(\mathcal{M}_{-k}) \), where $\mathcal{M}_k$ and $\mathcal{M}_{-k}$ represent the survival models with and without client $k$'s updates, respectively \cite{behera2023p2p}; \(\Omega\) is the model accuracy (C-index in survival analysis as discussed in Section \ref{sec:CI}).

\subsection{Aggregation of Feedback}
Based on the definition of reputation at the beginning of Section \ref{sec:repute}, the scores are updated based on peer feedback as:

\begin{equation*}
RS_{ik}(t+1) = RS_{ik}(t) + \alpha \sum_{j \in \text{Peers}_i} RS_{ij}(t) \cdot m_{j,k}
\end{equation*}

\noindent Here. $RS_{ij}(t)$ represents the reputation score assigned by client $i$ to client $j$, and $\alpha$ is the learning rate that governs the influence of peer feedback during reputation updates. This ensures feedback from reliable peers has a greater impact \cite{fang2024byzantine}.

\section{Proposed Framework for Reputation-Enhanced Federated Learning}

We use the Cox Proportional Hazards (CoxPH) model (see Section~\ref{sec:coxph}) within a federated learning (FL) framework, addressing data heterogeneity in patient counts and feature spaces across diverse medical centers. Each center contributes to a shared survival analysis model while preserving the privacy and integrity of its local data. To further safeguard sensitive information, we apply differential privacy (DP) to local model updates before sharing them for peer-based reputation assessment. This reputation is dynamically estimated based on the impact of each client's contributions on peer performance and is used to weigh model updates during aggregation. As illustrated in Figure~\ref{fig:schematic}, the approach decouples the reputation estimation from the aggregation process: DP-protected updates are shared among peers for reputation computation, while original updates are sent to the server for model aggregation. Clients then perform local updates using the globally aggregated model, modulated by their peer-evaluated reputation. 

\subsection{\textbf{Parameter Estimation in CoxPH Model}}
\label{subsec:beta-calculation-coxph}
The coefficient vector \(\beta\) is estimated by maximizing the partial likelihood over a dataset of patient features and potentially censored times of events. For any given dataset of \(p\) individuals, $X = \{x_i | 1\le i \le p\}$, the partial likelihood is:
\begin{equation}\label{eq:likely}
    L(\beta\mid X) = \prod_{i: \delta_i = 1} \left[ \frac{\exp(\beta^T x_i)}{\sum_{j \in R(t_i)} \exp(\beta^T x_j)} \right],
\end{equation}

Here, \(x_i\) is the covariate vector for the \(i\)-th individual, \(\delta_i\) is the event indicator (1 if the event occurred, 0 otherwise), \(R(t_i)\) is the set of individuals who have not experienced the event or censoring before time $t_i$ (also called risk set at time \(t_i\)), and \(\beta\) is the coefficient vector. 
We maximize the likelihood using iterative optimization (such as Newton-Raphson), where at each iteration, the \(\beta\) values are updated (using the \texttt{sksurv} package~\cite{sksurv}) to increase the partial likelihood.

We employ Breslow's method~\cite{breslow1975analysis} to estimate the baseline hazard $h_0(t)$ in \eqref{eq:CoxPH_hazard}. The estimated baseline hazard at each distinct failure time \(t_i\) is given by \( \hat{h}_0(t_i) = \frac{d_i}{\sum_{j \in R(t_i)} \exp(\beta^T x_j)} \), where \(d_i\) is the number of events at time \(t_i\). In this work, we represent the estimated survival model coefficients \(\beta\) collectively as \(\theta\), which are shared with and aggregated by the central server. Next, Algorithm~\ref{alg:reputation_fl_clustering} outlines the overall workflow for reputation-enhanced FL, where nodes exchange reputation score messages, assess peer reliability, and use these reputation scores to weight contributions during model aggregation.

\begin{algorithm}[ht]
\caption{Reputation-enhanced Federated Learning with Clustering-Based Noise Handling}
\label{alg:reputation_fl_clustering}
\KwIn{Initial reputation scores \(RS_{ij}(0)\); noise parameters \((T^{\text{honest}},T^{\text{ramp}},\epsilon_{\max})\); rounds \(R\); local datasets \(\{D_i\}\); clustering parameter \(\lambda\).}
\KwOut{Cluster-specific global survival analysis models \(\{\mathcal{M}_{\text{global}}^c\}\).}
\textbf{Init:}\\
Set \(R_{ij}(0)=1.0\) for all; assign noise vectors \(z_i\) with \(\sum_i z_i=0\).\\
Compute feature completeness vectors \(B_i\) and cluster clients into \(C_1,\ldots,C_c\) (minimizing Euclidean distance and maximizing C-Index).\\[1mm]
\For{\(r=1\) \KwTo \(R\)}{
  \tcp{Feedback \& reputation Update}
  \ForEach{client \(i\)}{
    \(\displaystyle RS_{ik}(r+1) \gets RS_{ik}(r)+\alpha \sum_{j\in\mathcal{P}_i}RS_{ij}(r)\,m_{j,k}\) \tcp*[r]{for all peers \(k\)}
  }
  \tcp{Client Selection per Cluster}
  \ForEach{cluster \(C_k\)}{
    Compute selection probability \(P(i)=\frac{RS_i(r)}{\sum_{j\in C_k}RS_j(r)}\) for each \(i\in C_k\)
  }
  \tcp{Local Training \& Aggregation}
  \ForEach{cluster \(C_k\)}{
    \ForEach{selected client \(i\in C_k\)}{
      Train local model \(\mathcal{M}_i\) on \(D_i\)
    }
    Aggregate: \(\displaystyle \mathcal{M}_{\text{global}}^k \gets \sum_{i\in C_k}P(i)\,\mathcal{M}_i\)
  }
  \tcp{Evaluation}
  \ForEach{client \(i\) in each \(C_k\)}{
    Compute the concordance index \(C_i\) and update reputation scores accordingly.
  }
}
\Return \(\{\mathcal{M}_{\text{global}}^c\}\).
\end{algorithm} 

\subsection{Initial Reputation Calculation}
Before the FL process begins, a message exchange phase enables nodes to evaluate the reliability of peers through reputation feedback and performance evaluation.
\begin{itemize}
    \item \textit{Message Exchange.} Nodes exchange performance metrics (e.g., local C-index scores) and provide feedback on their peers' contributions.
    \item \textit{Reputation Score Updates}. Each node dynamically updates its reputation scores based on feedback received from the peers.
    \item \textit{Feedback Definition.} A node $i$'s, feedback from node $j$ regarding node $k$'s contribution is: \( m_{j,k} = \omega(\mathcal{M}_k) - \omega(\mathcal{M}_{-k}) \), where $\mathcal{M}_k$ and $\mathcal{M}_{-k}$ is the survival analysis models with and without $k$'s updates, respectively \cite{behera2023p2p}.
\end{itemize}

\subsection{Aggregation via Feedback and Clustering}

Instead of excluding client nodes based on a fixed reputation threshold, our framework continuously updates reputation scores through decentralized peer feedback. Specifically, each node evaluates the contribution of its peers by measuring the improvement in local predictive performance -- quantified, for instance, by the change in the \textit{concordance index} (see Section \ref{sec:CI}) when incorporating the update of a peer's model. These performance improvements are used to adjust reputation scores in a gradual, iterative manner. In addition, we employ a custom clustering algorithm that groups nodes based on the completeness of their feature vectors and their predicted risk. Within each cluster, an aggregated reputation score is computed as the mean of peer evaluations. This decentralized mechanism inherently rewards nodes that provide high-quality updates and diminishes the influence of nodes whose contributions are deemed unreliable or adversarial. Notably, by decoupling reputation updates from federated model aggregation, our approach ensures that clients do not gain knowledge of their peers' data or local model parameters, preserving privacy while still mitigating unreliable or adversarial contributions

\begin{figure*}[t!]
    \centering    
    \includegraphics[width=0.8\textwidth]{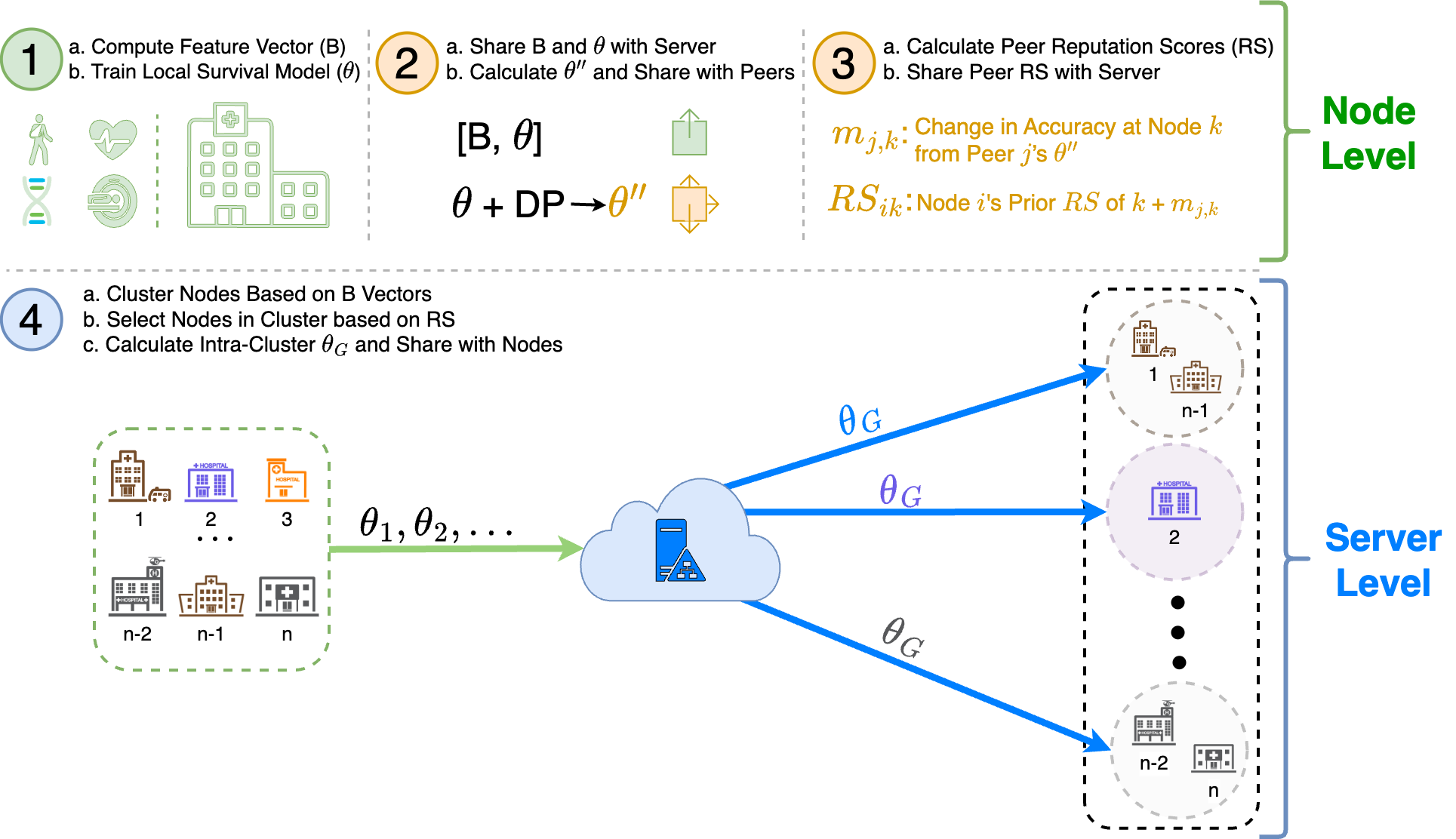}
    \caption{The detailed workflow of the proposed reputation-enhanced federated survival analysis framework, showing node-level steps (1–3) for B-vector computation, local CoxPH training, DP-protected reputation scoring, and server-level clustering and global model aggregation (4)}
    \label{fig:over}
\end{figure*}

\subsection{\textbf{Reputation-Enhanced FL Framework}}
\label{subsec:reputation-enhanced-fl-framework}
This builds upon the standard FL approach by incorporating a decentralized, peer-driven reputation mechanism and clustering to improve robustness against noisy or malicious clients.  

\noindent \textit{1. Initialization and Clustering.} Each client \(i\) computes a feature completeness vector \(B_i\), where each element represents the fraction of non-null values for a global feature. The clients are clustered into \(c\) clusters \(\{C_1, C_2, ..., C_c\}\) based on the similarity of their feature completeness vectors \(B_i\) and predicted risk scores, minimizing the objective function defined in Section \ref{sec:form}.  This ensures that clients within a cluster have similar data characteristics. The reputation that client \( i \) for peer \( j\), i.e., \( T_{ij}(0) \), is initialized to \(1.0\).

\textit{2. Local Calculations and Parameter Estimation (within each round \(r\)).} Each client \(i\) in cluster \(C_k\) trains a local CoxPH model on its private dataset \(D_i\) using the current global model parameters \(\beta^{(t)}\) from the previous round (or initialization). The local model update is performed on the parameters of the local CoxPH model, \(\theta_i(t)\). Therefore, the local optimization aims to minimize the local loss function \(\mathcal{L}_i(\beta)\), given by:

    \[
    \theta_i(t) = \arg\min_{\theta} \mathcal{L}_i(\theta; \beta^{(t)})
    \]

\textit{3. Reputation-Weighted Message Exchange and Aggregation (within each cluster \(C_k\))}. Within each cluster \(C_k\), each client \( i \) is selected for participation in the aggregation step with a probability proportional to their reputation scores, given by:
    \[
    P(i) = \frac{RS_i(t)}{\sum_{j \in C_k} RS_j(t)}
    \]
\noindent The selected clients exchange their local model updates \(\theta_i(t)\) as well as reputation feedback messages.  For client \(i\), the feedback from client \(j\) regarding client \(k\)'s contribution is measured by: \( m_{j,k}(t) = \Omega(\mathcal{M}(\theta_j(t),\theta_k(t))) - \Omega(\mathcal{M}(\theta_{j}(t))) \), where, \(\mathcal{M}(\theta_j(t),\theta_k(t))\) is a survival analysis model evaluated using local parameters and client \(k\)'s update, and \(\mathcal{M}(\theta_{j}(t))\) is a survival analysis model evaluated without client \(k\)'s update. The accuracy \( \Omega \) is measured in terms of the concordance index (refer to Section \ref{sec:CI}).  This feedback reflects the impact of client \(k\)'s update on client \(j\)'s local model performance.

\textbf{Differential Privacy (DP) for Secure Reputation Computation}. To protect sensitive information during peer-to-peer sharing for reputation score calculation, each client applies DP to their local update \(\theta_i(t)\) before sharing with peers. The update is a vector of CoxPH coefficients, where index \(j\) in \(\theta_{i,j}(t)\) denotes the \(j\)-th component (corresponding to the \(j\)-th covariate, \(j = 1, \dots, d\)). The update is clipped to have an L2 norm at most \(Q\), i.e., \(\theta_i'(t) = \theta_i(t) \cdot \min(1, \frac{Q}{\|\theta_i(t)\|_2})\), where \(\|\theta_i(t)\|_2 = \sqrt{\sum_{j=1}^d \theta_{i,j}(t)^2}\). Then, Gaussian noise is added: \(\theta_i''(t) = \theta_i'(t) + \mathcal{N}(0, \sigma^2 I)\), with \(\sigma = \frac{Q \sqrt{2 \ln(1.25 / \delta)}}{\zeta}\), ensuring \((\zeta, \delta)\)-differential privacy \cite{dwork2006calibrating,abadi2016deep}. The noise scale \(\sigma\) is chosen to satisfy the Gaussian mechanism’s requirement for \((\zeta, \delta)\)-DP, balancing privacy (via \(\zeta, \delta\)) and accuracy (via \(Q\)). Here, \(\zeta\) is a real positive value. The noisy update \(\theta_i''(t)\) is shared with peers, while the non-private update \(\theta_i(t)\) is sent to the server for aggregation, mitigating privacy risks in peer interactions and minimally affecting global model updates.

\textbf{Privacy and Accuracy Guarantee.} The mechanism ensures privacy and preserves accuracy with appropriate \(Q\), \(\zeta\), and \(\delta\) \cite{dwork2006calibrating,mcmahan2017learning}.

See \ref{sec:dp_proof} for a detailed proof and accuracy guarantees. Finally, each client \(i\) updates its reputation scores based on the received feedback:
    \begin{equation}
        \label{eq:reputation_update}
    RS_{ik}(t+1) = RS_{ik}(t) + \alpha \sum_{j \in \text{selected peers}} RS_{ij}(t) \cdot m_{j,k}(t)
    \end{equation}

\noindent Here, \(\alpha\) is the learning rate for the reputation score update, which controls neighbor importance during reputation updates.

\noindent The global model parameters for cluster \(C_k\) are updated by estimating the reputation-weighted average of the local updates of the selected clients. It is given by:
\[ \beta^{(t+1)} =  \sum_{i \in C_k} P(i) \theta_i(t) = \frac{\sum_{i \in C_k} RS_i(t) \theta_i(t)}{\sum_{i \in C_k} RS_i(t)} 
\]

\textbf{4. Global Model Dissemination.} The updated global model parameters \(\beta^{(t+1)}\) for each cluster \(C_k\) are sent to the clients assigned to that cluster. Steps 2-4 are repeated for a predefined number of rounds \(R\) to ensure that the reputation scores upon convergence reflect the behavior of the participating clients.

Figure~\ref{fig:over} depicts the detailed workflow of the reputation-enhanced federated survival analysis framework.

\subsection{Convergence Analysis of Peer-Driven Reputation Estimation}
\noindent \textbf{Lemma.} \textit{Given reliable feedback, the reputation estimated by a client converges to the true reliability of the peer after sufficient iterations of message passing.}\\
\noindent \textbf{Case 1.} Updating reputation score in a three-node setting:\\

\begin{proof}
Let \( RS_{uv}(t) \) be the reputation estimation of node \( u \) towards node \( v \) at time \( t \), and \( RS_v^* \) be the true reliability of node \( v \). We define the error \( e_v(t) \) as the difference between the true reliability and the estimated reputation, i.e., \( e_v(t) = RS_v^* - RS_{uv}(t) \).

Given the reputation update rule: \( RS_{uv}(t + 1) = RS_{uv}(t) + \alpha \times RS_{uw}(t) \times m_{wv}(t) \), where \( m_{wv}(t) = RS_v^* + \eta \), and \( \eta \) is a zero-mean noise term with finite variance, the error at time \( t + 1 \) is given by: \( e_v(t + 1) = RS_v^* - \left[ RS_{uv}(t) + \alpha \times RS_{uw}(t) \times m_{wv}(t) \right] \). Thus, we have:
\[
e_v(t + 1) = e_v(t) - \alpha \times RS_{uw}(t) \times [RS_v^* + \eta].
\]
\noindent Taking the expectation of both sides: 
\[ \mathbb{E}[e_v(t + 1)] = \mathbb{E}[e_v(t)] - \alpha \times \mathbb{E}[RS_{uw}(t)] \times \mathbb{E}[RS_v^*] \]. 
\noindent Since \( \mathbb{E}[\eta] = 0 \), we simplify to: \( \mathbb{E}[e_v(t + 1)] = \mathbb{E}[e_v(t)] - \alpha \times \mathbb{E}[RS_{uw}(t)] \times RS_v^*\). Now, observe that the second term, \( \alpha \times \mathbb{E}[RS_{uw}(t)] \times RS_v^* \), is a constant concerning the error term \( e_v(t) \). Thus, \( \mathbb{E}[e_v(t + 1)] = (1 - \alpha \times \mathbb{E}[RS_{uw}(t)] \times RS_v^*) \times \mathbb{E}[e_v(t)]  \). As \( \alpha \) is small enough, this term will reduce the error term by a factor that remains between 0 and 1, leading to:
\[
\mathbb{E}[e_v(t)] \to 0 \quad \text{as} \quad t \to \infty.
\]
The estimated reputation \( RS_{uv}(t) \) converges to the true reputation score \( RS_v^* \) as the error becomes negligible over time, provided the feedback mechanism is ideal.
\end{proof}

\noindent \textbf{Case 2.} Updating reputation score in a \( n \)-node setting:
\begin{proof} The reputation estimation update now incorporates feedback from \( n \) clients. The reputation update rule is:
\[
RS_{uv}(t+1) = RS_{uv}(t) + \alpha \sum_{w \in \{w_1, w_2, \dots, w_n\}} RS_{uw}(t) \times m_{wv}(t),
\]At time \( t+1 \), the error evolves as: \( e_v(t+1) = RS_v^* - \left[ RS_{uv}(t) + \alpha \sum_{w \in \{w_1, w_2, \dots, w_n\}} RS_{uw}(t) \times m_{wv}(t) \right] \).
Expanding this, by plugging \( e_v(t) = RS_v^* - RS_{uv}(t) \)
\[
e_v(t+1) = e_v(t) - \alpha \sum_{w \in \{w_1, w_2, \dots, w_n\}} RS_{uw}(t) \times [RS_v^* + \eta_w(t)].
\]
Since the second term is once again a constant w.r.t the error and \( \mathbb{E}[\eta_w(t)] = 0 \), we simplify to: \( \mathbb{E}[e_v(t+1)] = \left(1 - \alpha \sum_{w \in \{w_1, w_2, \dots, w_n\}} \mathbb{E}[RS_{uw}(t)] \times RS_v^* \right) \mathbb{E}[e_v(t)] \).

Since the coefficient of \( \mathbb{E}[e_v(t)] \) remains between 0 and 1 for a sufficiently small \( \alpha \), it follows that: \( \mathbb{E}[e_v(t)] \to 0 \quad \text{as} \quad t \to \infty. \). Thus, in the presence of feedback from multiple clients, the estimated reputation \( RS_{uv}(t) \) still converges to the true reliability \( RS_v^* \).
\end{proof}

\subsection{Differential Privacy: Proof of Accuracy Guarantees}
\label{sec:dp_proof}

We prove that differential privacy (DP) applied to local updates \(\theta_i(t)\) during peer-to-peer reputation score calculation preserves accuracy in reputation scores \(RS_{ik}(t)\), ensuring effective client selection and global model performance.

\subsubsection{Assumptions}
\begin{itemize}
    \item Client \(i\) computes \(\theta_i(t) = \arg\min_{\theta} \mathcal{L}_i(\theta; \beta^{(t)})\), where \(\mathcal{L}_i\) is the CoxPH loss on dataset \(D_i\), \(\beta^{(t)}\) is the global model at round \(t\).
    \item Reputation scores update: \(RS_{ik}(t+1) = RS_{ik}(t) + \alpha \sum_{j \in \text{selected peers}} RS_{ij}(t) \cdot m_{j,k}(t)\), with \(m_{j,k}(t) = \Omega(\mathcal{M}(\theta_j(t),\theta_k(t))) - \Omega(\mathcal{M}(\theta_{j}(t)))\) and $\Omega$ being the concordance index (Section \ref{sec:CI}).
    \item DP clips \(\theta_i(t)\) to norm \(Q\): \(\theta_i'(t) = \theta_i(t) \cdot \min(1, \frac{Q}{\|\theta_i(t)\|_2})\), adds noise: \(\theta_i''(t) = \theta_i'(t) + \mathcal{N}(0, \sigma^2 I)\), \(\sigma = \frac{Q \sqrt{2 \ln(1.25 / \delta)}}{\zeta}\).
    \item \(\Omega\) is \(L_\Omega\)-Lipschitz; \(RS_{ij}(t) \in [0, T_{\max}]\); \(N_p\) peers; parameter dimension \(d\); global loss \(\mathcal{L}_{C_k}(\beta) = \frac{\sum_{i \in C_k} RS_i(t) \mathcal{L}_i(\beta)}{\sum_{i \in C_k} RS_i(t)}\) is \(L_k\)-smooth.
\end{itemize}

\subsubsection{Accuracy Guarantee}
Reputation scores use DP updates \(\theta_i''(t)\), so errors may affect selection probabilities \(P(i) = \frac{RS_i(t)}{\sum_{j \in C_k} RS_j(t)}\), reducing high-quality client influence. We bound these errors to ensure effective selection and model performance~\cite{geyer2017differentially}.

\textbf{Reputation Feedback Error}: For feedback \(m_{j,k}(t) = \Omega(\mathcal{M}(\theta_j(t),\theta_k(t))) - \Omega(\mathcal{M}(\theta_{j}(t)))\), with \(\theta_k''(t) = \theta_k'(t) + \eta_k\), \(\eta_k \sim \mathcal{N}(0, \sigma^2 I)\), and \(\Omega\) being \(L_\Omega\)-Lipschitz, the error is:
    \[
    |m_{j,k}''(t) - m_{j,k}(t)| \leq L_\Omega \|\eta_k\|_2,\]
    \[ \quad \mathbb{E}[\|\eta_k\|_2] \leq \sqrt{d} \cdot \frac{Q \sqrt{2 \ln(1.25 / \delta)}}{\zeta},
    \]
    \[
    \mathbb{E}[|m_{j,k}''(t) - m_{j,k}(t)|] \leq L_\Omega \sqrt{d} \cdot \frac{Q \sqrt{2 \ln(1.25 / \delta)}}{\zeta}.
    \]
    This bounds the expected error in reputation feedback due to DP noise, proving the feedback deviation is controlled by \(Q\) and \(\zeta\). The error is thus bounded by a real scalar determined by the individual values of \(L_\Omega\), \(d\), \(Q\), \(\delta\), and \(\zeta\), ensuring a predictable impact of DP noise on reputation feedback.

\textbf{Reputation Score Error}: With \(RS_{ik}(t+1) = RS_{ik}(t) + \alpha \sum_{j \in \text{selected peers}} RS_{ij}(t) \cdot m_{j,k}(t)\), the error is:
    \[
    \mathbb{E}[|RS_{ik}''(t+1) - RS_{ik}(t+1)|] \]
    \[
    \leq \alpha N_p T_{\max} L_\Omega \sqrt{d} \cdot \frac{Q \sqrt{2 \ln(1.25 / \delta)}}{\zeta}.
    \]
    This shows the expected error in reputation scores per round is proportional to DP noise, proving scores remain close to true values with small \(Q\) and large \(\zeta\) as real positive quantities. The error is thus bounded by a real scalar determined by the individual values of \(\alpha\), \(N_p\), \(T_{\max}\), \(L_\Omega\), \(d\), \(Q\), \(\delta\), and \(\zeta\), ensuring a predictable impact on reputation scores.

\textbf{Client Selection Impact}: For \(P(i) = \frac{RS_i(t)}{\sum_{j \in C_k} RS_j(t)}\), let \(RS_i(t) = RS_i^{\text{true}}(t) + \zeta_i(t)\), where \(\mathbb{E}[|\zeta_i(t)|] \leq \alpha N_p T_{\max} L_\Omega \sqrt{d} \cdot \frac{Q \sqrt{2 \ln(1.25 / \delta)}}{\zeta}\). With \(\sum_{j \in C_k} RS_j(t) \geq S_{\min} > 0\),
    \[
    \mathbb{E}[|P''(i) - P(i)|] \leq \]
    \[
    \frac{\alpha N_p T_{\max} L_\Omega \sqrt{d} \cdot \frac{Q \sqrt{2 \ln(1.25 / \delta)}}{\zeta}}{S_{\min}} \left(1 + \frac{|C_k| T_{\max}}{S_{\min}}\right).
    \]
    This bounds the error in selection probabilities, proving DP minimally affects client selection with proper \(Q\) and \(\zeta\). The error is thus bounded by a real scalar determined by the individual values of \(\alpha\), \(N_p\), \(T_{\max}\), \(L_\Omega\), \(d\), \(Q\), \(\delta\), \(\zeta\), \(S_{\min}\), and \(|C_k|\), ensuring a predictable impact on client selection probabilities.

\textbf{Global Model Impact}: For \(\beta^{(t+1)} = \sum_{i \in C_k} P(i) \theta_i(t)\), define \(\Theta_{\max} = \max_i \|\theta_i(t)\|_2\) (maximum L2 norm of updates). The error is:
    \[
    \mathbb{E}[\|\beta''^{(t+1)} - \beta^{(t+1)}\|_2] \leq |C_k| \Theta_{\max} \]
    \[
    \cdot \frac{\alpha N_p T_{\max} L_\Omega \sqrt{d} \cdot \frac{Q \sqrt{2 \ln(1.25 / \delta)}}{\zeta}}{S_{\min}} \left(1 + \frac{|C_k| T_{\max}}{S_{\min}}\right).
    \]
    Since \(\mathcal{L}_{C_k}\) is \(L_k\)-smooth, the loss increase is:
    \[
    \mathbb{E}[\mathcal{L}_{C_k}(\beta''^{(t+1)}) - \mathcal{L}_{C_k}(\beta^{(t+1)})] \leq \]
    \[ \frac{L_k}{2} \left( |C_k| \Theta_{\max} \frac{\alpha N_p T_{\max} L_\Omega \sqrt{d} \cdot \frac{Q \sqrt{2 \ln(1.25 / \delta)}}{\zeta}}{S_{\min}} \right. \]
    \[
    \left(1 + \frac{|C_k| T_{\max}}{S_{\min}}\right) \left. \right)^{2}.
    \]
 This bounds the expected loss increase in the global model, proving DP’s impact on model performance is minimal with tuned parameters. The loss increase is thus bounded by a real scalar determined by the individual values of \(L_k\), \(|C_k|\), \(\Theta_{\max}\), \(\alpha\), \(N_p\), \(T_{\max}\), \(L_\Omega\), \(d\), \(Q\), \(\delta\), \(\zeta\), and \(S_{\min}\), ensuring a predictable impact on the global model performance.

\section{Experimental Results}

We validate the proposed system on simulated and real datasets, comparing its performance against a baseline FL approach that incorporates reputation in model aggregation.

\subsection{Simulation Studies}
Our analysis, conducted across 10 centers with (\( r=5000 \)) patients. To model real-world variability, synthetic datasets were generated with 50 possible features per patient. Each center included 10 shared features across all centers, while the remaining features were randomly selected for each center to capture diverse clinical environments.

The features for each patient, \( \mathbf{x}^{(i)} \), were sampled from a normal distribution: \( \mathbf{x}^{(i)} \sim \mathcal{N}(0, I_p/p) \), where \( p \) represents the number of features, ensuring \( \mathbf{E}\left[\lVert \mathbf{x}^{(i)} \rVert^2 \right] = 1 \). Here, \( I_p \) denotes the identity matrix of size \( p \times p \), implying that each feature is independently and identically distributed with variance \( \frac{1}{p} \). To reflect real-world clinical settings, a controlled missingness mechanism was introduced. Each center exhibited a fraction \( \phi \) of features with missing values, where the missingness per feature was randomly selected within a predefined range, with the following attributes: (a) A fraction \( \alpha \) of features in each center had missing values. (b) The proportion of missing entries per affected feature followed a uniform distribution within a range \( [\rho_{\min}, \rho_{\max}] \), where \( \rho_{\min} \) and \( \rho_{\max} \) were varied across experiments. (c) Missing values were introduced using a \textit{missing completely at random} (MCAR), where the probability of a feature being missing was independent of both observed and unobserved data~\cite{little2019statistical}.

The event times \( \tau_i \) were modeled using a Cox proportional hazards (CoxPH) model with a baseline hazard \( h_0 \). Right censoring was applied to model incomplete follow-up, where censoring times were drawn from a uniform distribution: \( C_i \sim U(0, \Psi^{-1}(1/2))
\), where \( \Psi(u) \equiv \mathbb{P}[\tau < u] = 1 - e^{-u \exp (\beta^T \mathbf{x}_i)}\)~\cite{andreux2020federated}. The observed time for a patient was \( T_i = \min(\tau_i, C_i) \) with an event indicator \( \delta_i = \mathbb{I}(\tau_i \leq C_i) \).

To systematically evaluate the impact of missing data, we conducted experiments across multiple scenarios with varying: (a) feature missingness fractions \( \alpha \in \{0.1, 0.3, 0.5\} \); (b) missingness severity \( (\rho_{\min}, \rho_{\max}) \) set to (0,0.3), (0.3,0.6), and (0.6,0.99); and (c) censoring rates \( \gamma \in \{0.2, 0.4, 0.6\} \). 

\subsubsection{\textbf{Adaptive Noise Injection}}

We introduce a progressive noise injection strategy where nodes behave honestly in the initial phases to gain reputation before gradually introducing noise into their contributions after a predefined period. Please refer to Section \ref{sec:threat} on the proposed threat model.

\subsubsection{\textbf{Noise Injection Mechanism}}

To introduce realistic disruptions, noise is drawn from different probability distributions, representing varied adversarial interference. These include the \textit{Gaussian noise}~\cite{goodfellow2014explaining}, which is common in data poisoning attacks; \textit{uniform noise}~\cite{liu2017delving}, representing random corruptions in feature vectors; \textit{Poisson noise}~\cite{foi2008practical}, which generates sporadic spikes in feature values; \textit{Laplace noise}~\cite{dwork2006calibrating}, often used in differential privacy settings; \textit{speckle (salt-and-pepper) noise}~\cite{gonzalez2008digital}, which simulates sensor-induced distortions; and \textit{Cauchy noise}~\cite{huber2011robust}, representing long-tailed adversarial effects. For different types, the noise is injected into either model parameters, reputation messages, or both, allowing an analysis of attack vectors and their impact on reputation-based FL.

\subsubsection{\textbf{Impact of Noise on reputation Evolution}}

\begin{figure*}[ht]
    \centering
    \begin{tabular}{cc}
         \includegraphics[width=0.48\linewidth]{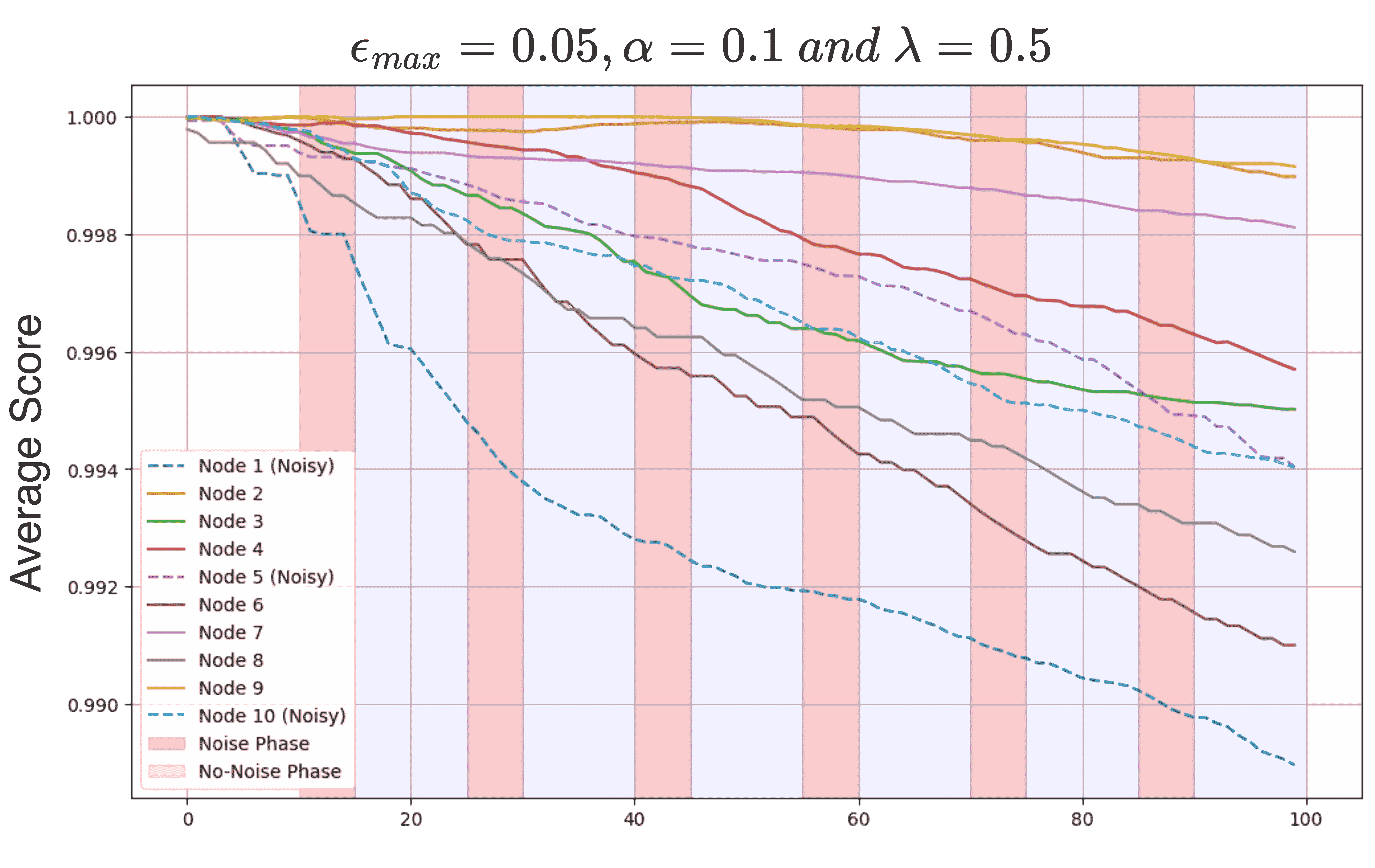}
         &
         \includegraphics[width=0.48\linewidth]{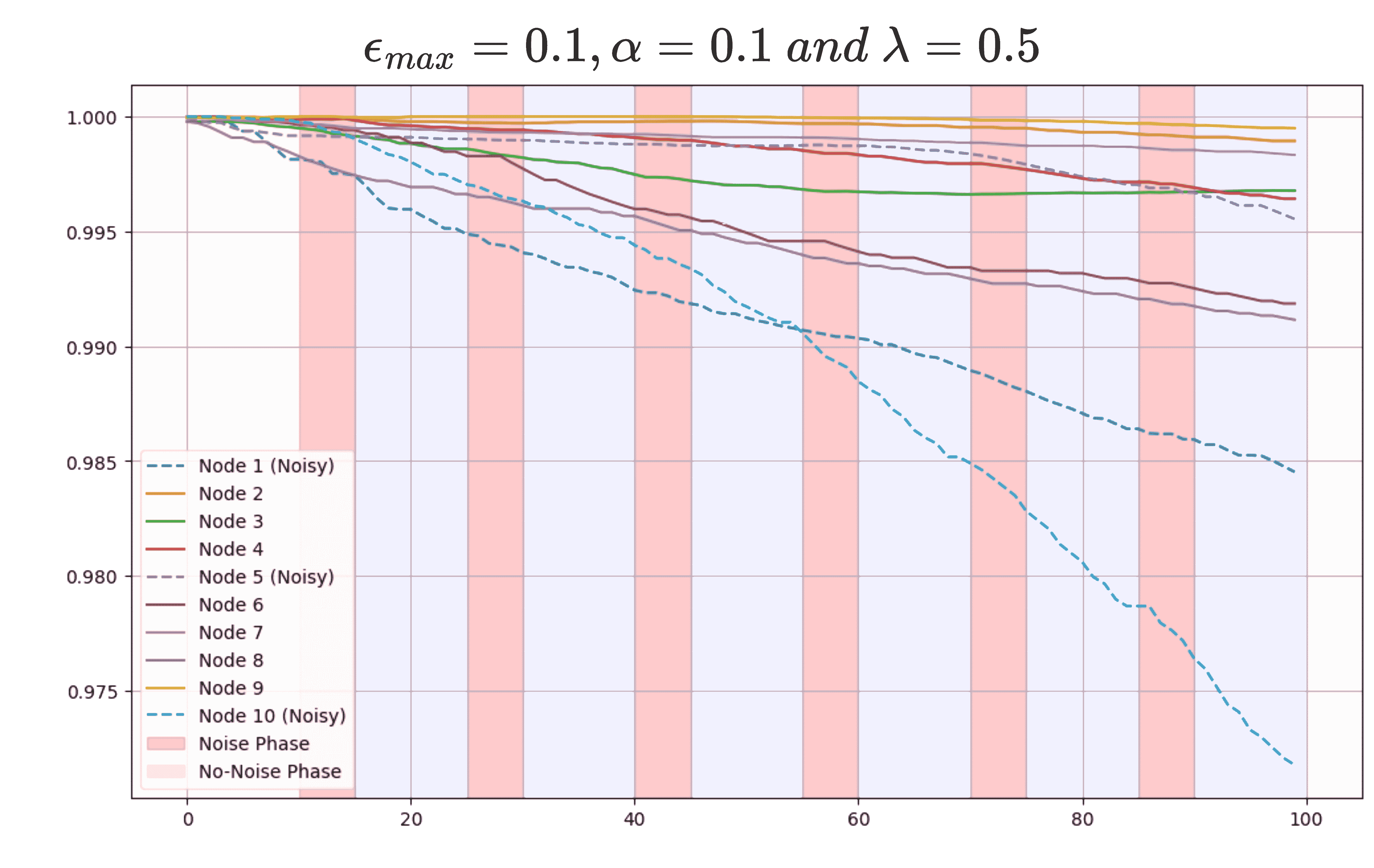}
         \\
         \includegraphics[width=0.48\linewidth]{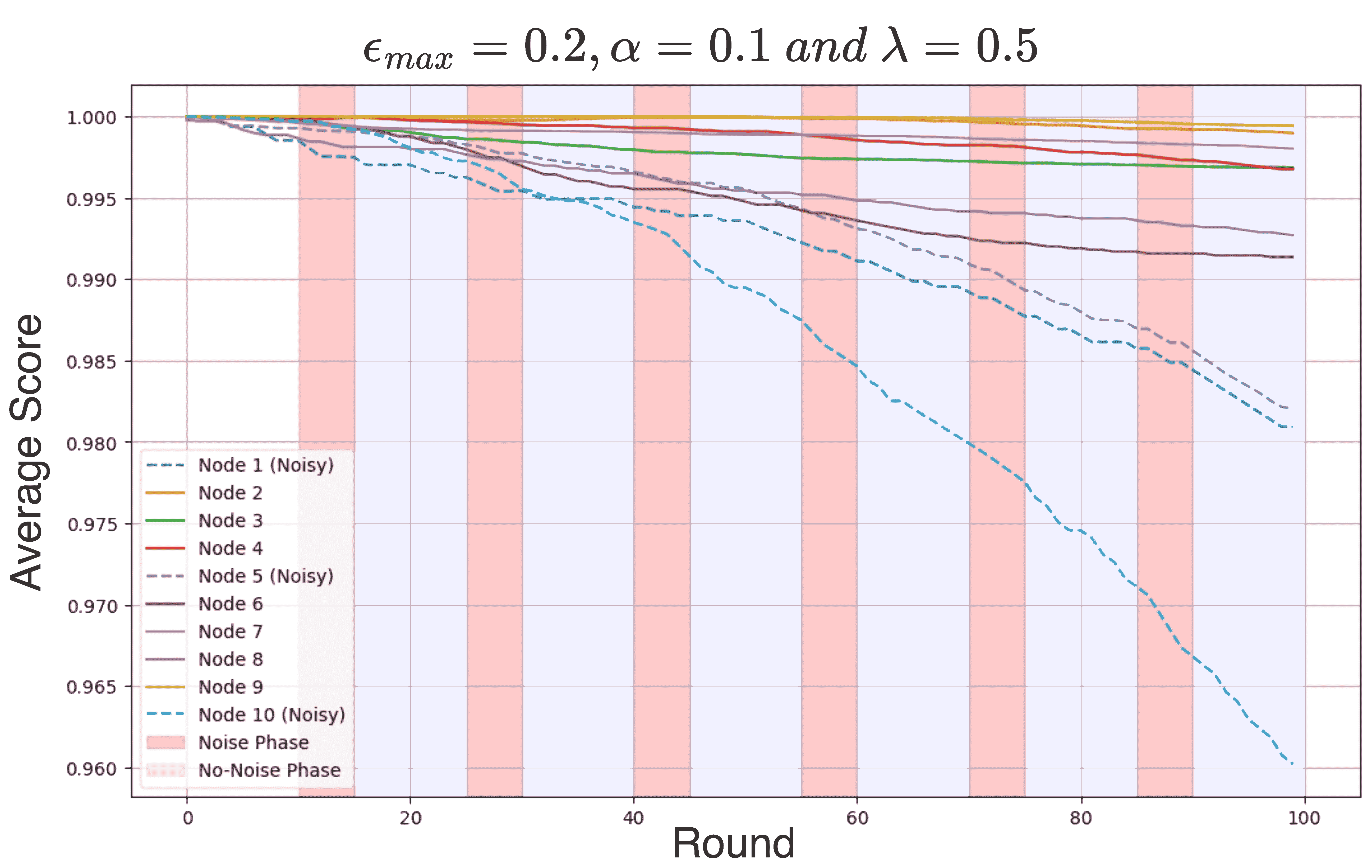}
         &
         \includegraphics[width=0.48\linewidth]{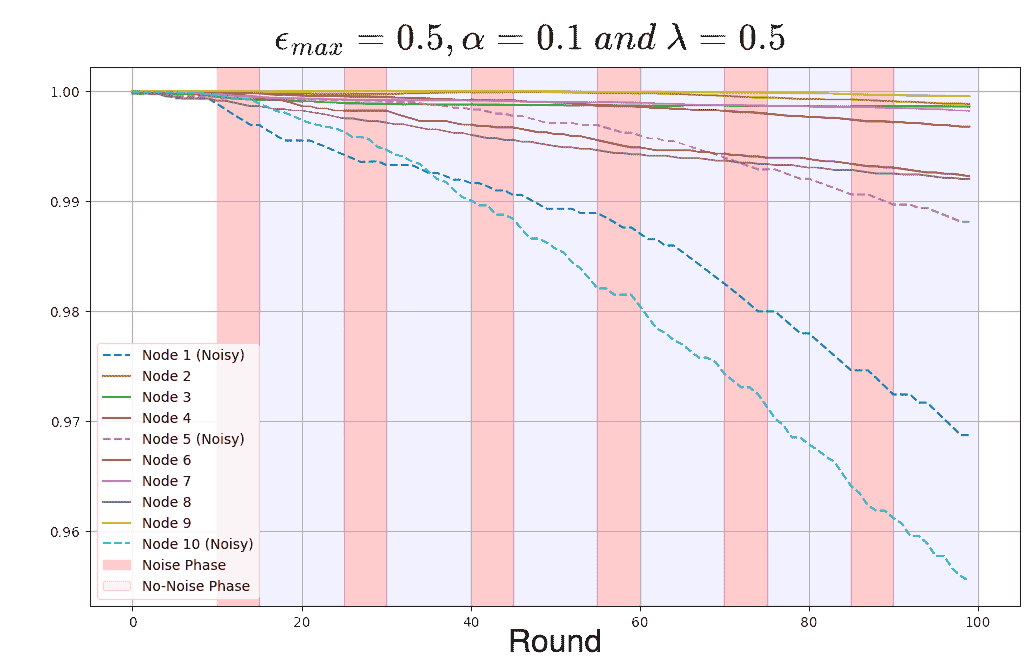}
    \end{tabular}
    
    \caption{Reputation score evolution over rounds for different noise levels (\(\epsilon_{\max}\)).
    The three plots correspond to \(\epsilon_{\max} = 0.05\), \(\epsilon_{\max} = 0.1\), \(\epsilon_{\max} = 0.2\), and \(\epsilon_{\max} = 0.5\), respectively, where \( \epsilon_{\max} \) corresponds to the maximum injectable noise. Each line represents a node's average reputation score, with noisy nodes shown using dashed lines.}
    \label{fig:reputation_evolution}
\end{figure*}

We analyze the impact of varying levels of injected noise on reputation evolution in federated learning. The parameter \(\epsilon_{\max}\) controls the maximum noise magnitude, and its effect is visualized across three values: \(\epsilon_{\max} = 0.05\), \(\epsilon_{\max} = 0.1\), \(\epsilon_{\max} = 0.2\), and \(\epsilon_{\max} = 0.5\). The results are depicted in Figure~\ref{fig:reputation_evolution}, where the average node reputation score is tracked over multiple rounds. In the low noise scenario (\(\epsilon_{\max} = 0.05\)), noisy nodes maintain relatively high reputation scores for an extended period; although their reputation gradually declines over rounds, the reduction is slow, indicating that more rounds are needed for the model to effectively detect and down-weight these noisy participants. In the moderate noise scenario (\(\epsilon_{\max} = 0.1\)), noisy nodes experience a steeper decline in reputation scores, demonstrating that the model efficiently distinguishes them from reliable nodes and rapidly reduces their influence, thereby improving robustness against adversarial contributions. In the high noise scenarios (\(\epsilon_{\max} = 0.2\) and \(\epsilon_{\max} = 0.5\)), noisy nodes are detected much earlier, with their reputation scores decreasing significantly within the initial rounds; concurrently, reliable nodes maintain high reputation levels, ensuring that the model effectively isolates unreliable participants and achieves robust aggregation.
The experiment demonstrates that as \(\epsilon_{\max}\) increases, the model more efficiently detects and suppresses noisy nodes by decreasing their reputation scores. It is worth noting that the proposed reputation-weighted federated averaging and clustering approach (1) dynamically adjusts node participation based on reliability and (2) exhibits the discriminative potential to adaptively identify low-reputation behavior under highly noisy environments.

\begin{figure*}[ht]
    \centering
    \begin{tabular}{cc}
         \includegraphics[width=0.48\linewidth]{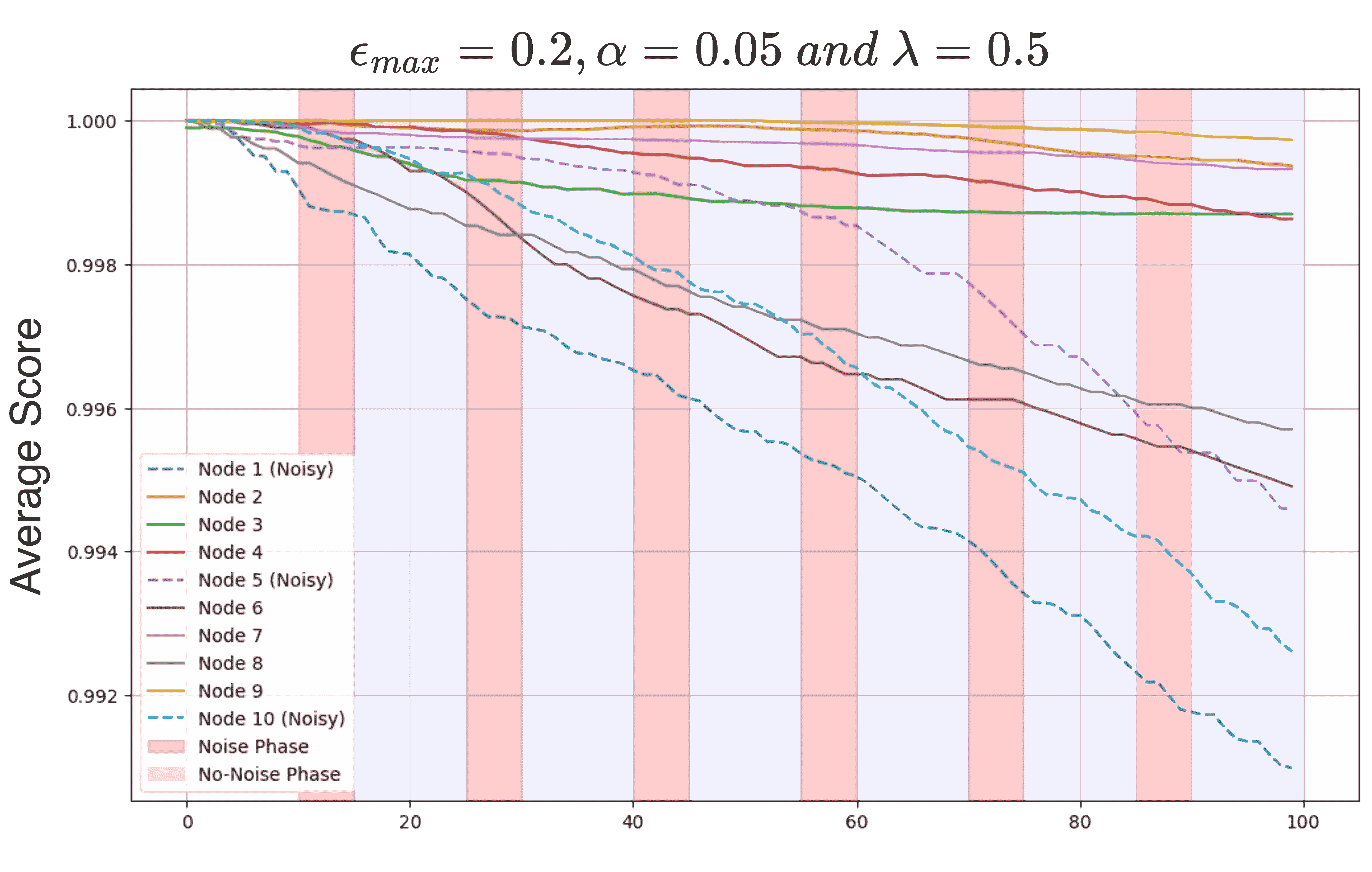}
         &
         \includegraphics[width=0.48\linewidth]{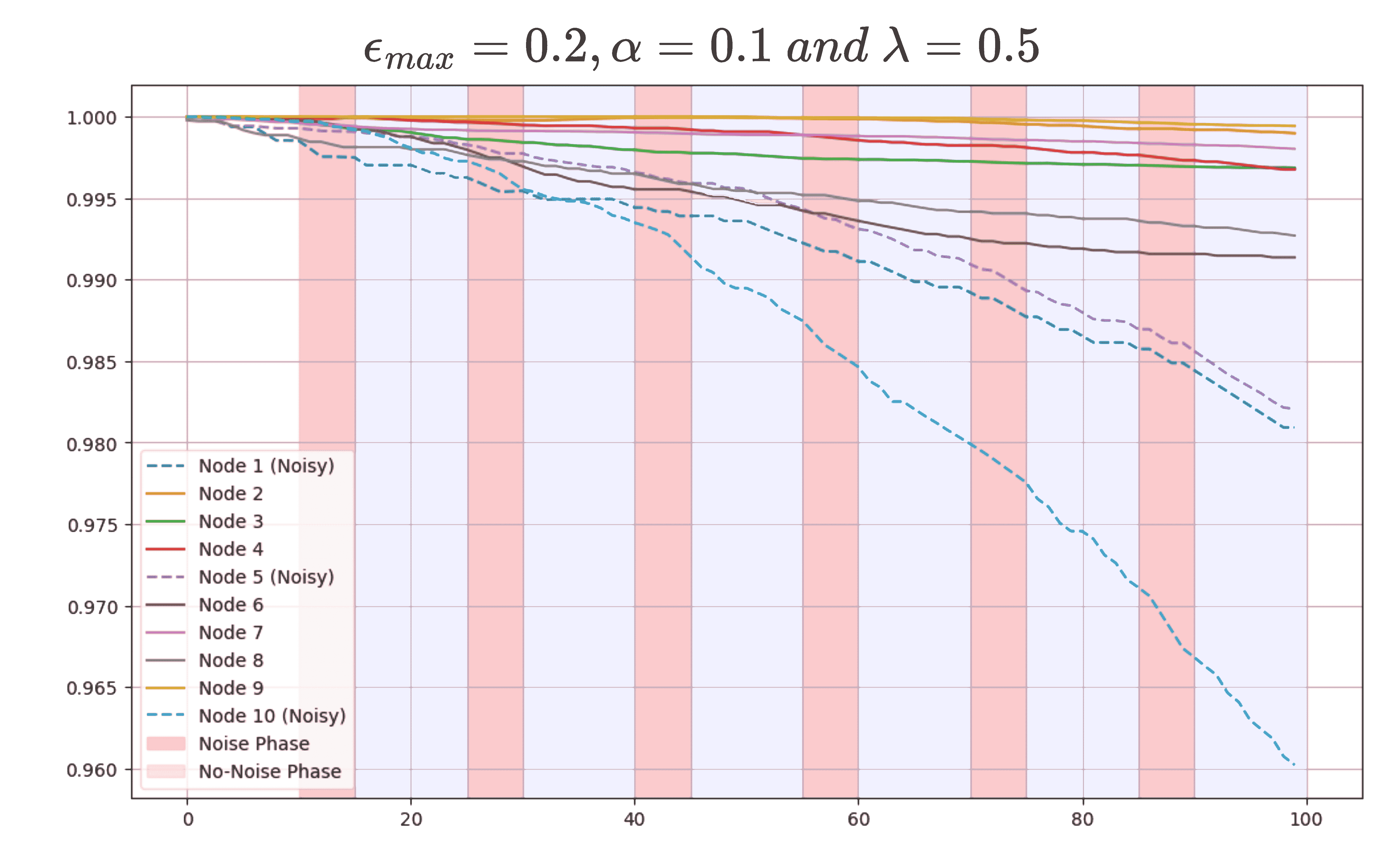}
         \\
         \includegraphics[width=0.48\linewidth]{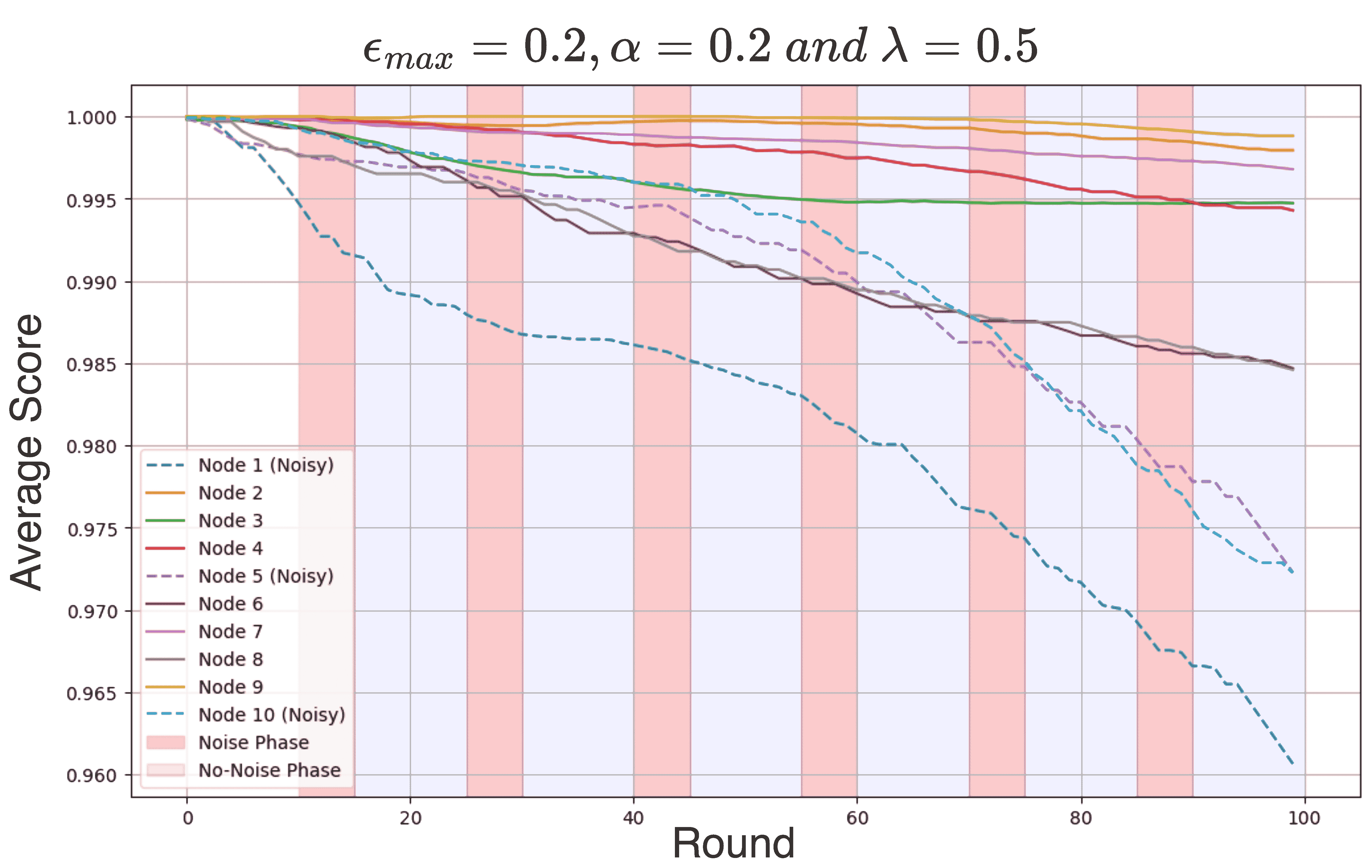}
         &
         \includegraphics[width=0.48\linewidth]{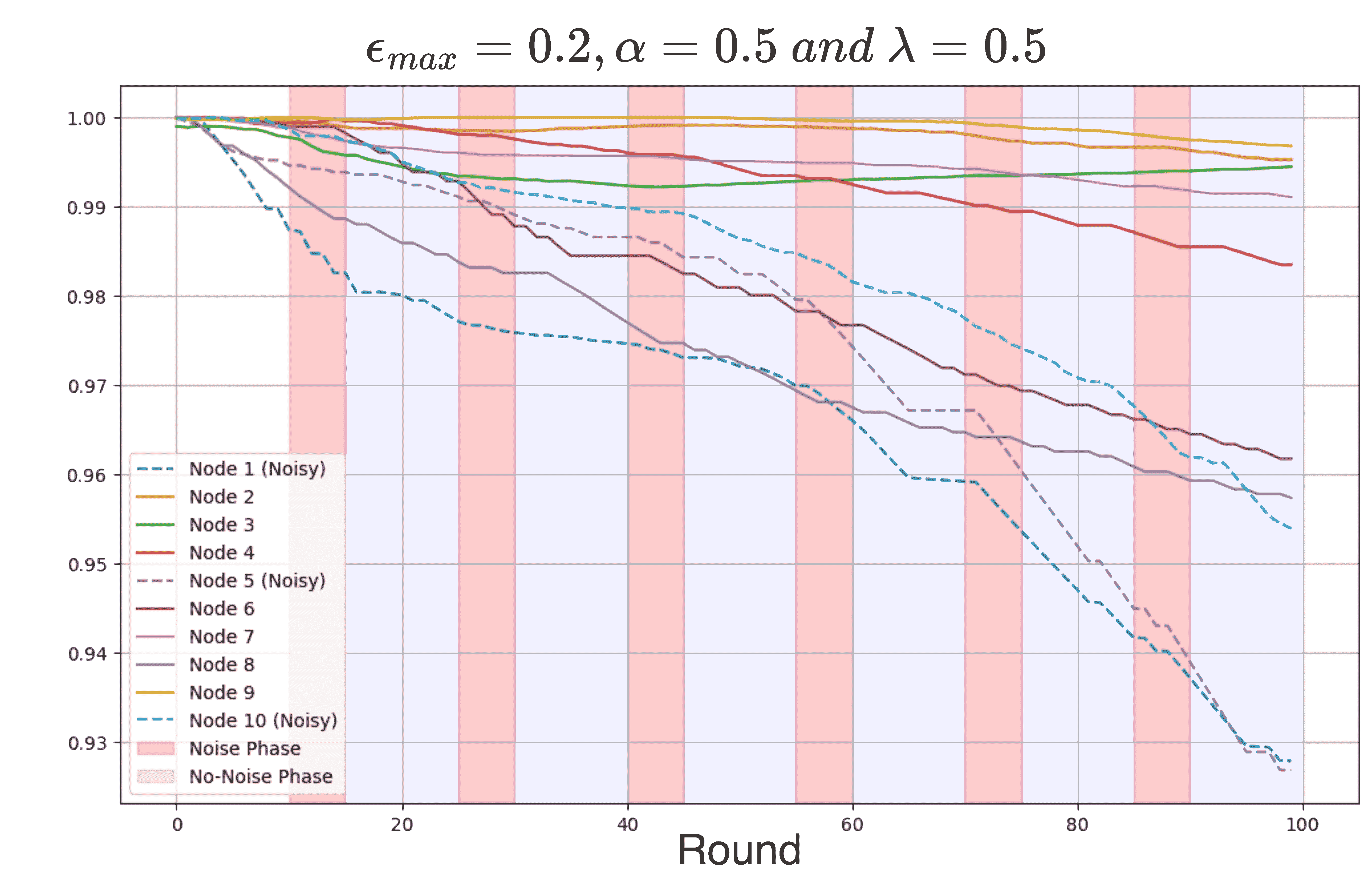}
    \end{tabular}

    \caption{Reputation score evolution over rounds for different values of neighbor importance during reputation updates (\(\alpha\)). The four plots correspond to \(\alpha = 0.05\), \(\alpha = 0.1\), \(\alpha = 0.2\), and \(\alpha = 0.5\), respectively.}
    \label{fig:alpha_reputation}
\end{figure*}

\subsubsection{\textbf{Impact of \(\alpha\) on reputation Score Evolution}}

Recall from our discussion in Section \ref{subsec:reputation-enhanced-fl-framework}, the parameter \(\alpha\) controls neighbor importance during reputation updates (\ref{eq:reputation_update}). The experimental results in Figure~\ref{fig:alpha_reputation} show that in the low \(\alpha\) scenario (e.g., \(0.05\)), the model updates reputation scores conservatively, resulting in a gradual separation between noisy and normal nodes, with noisy nodes retaining relatively high scores over an extended period before being penalized. In the moderate \(\alpha\) scenario (e.g., \(0.1\)), the model more effectively detects and penalizes noisy nodes while maintaining stable reputation scores for normal nodes, balancing robustness and adaptability. In contrast, for high \(\alpha\) values (\(0.2\) and \(0.5\)), the model reacts very aggressively to deviations, leading to a rapid decline in reputation of noisy nodes; however, such high sensitivity also risks over-penalizing nodes with only minor fluctuations.

\begin{figure*}[ht]
    \centering 
    \begin{tabular}{cc}
         \includegraphics[width=0.48\linewidth]{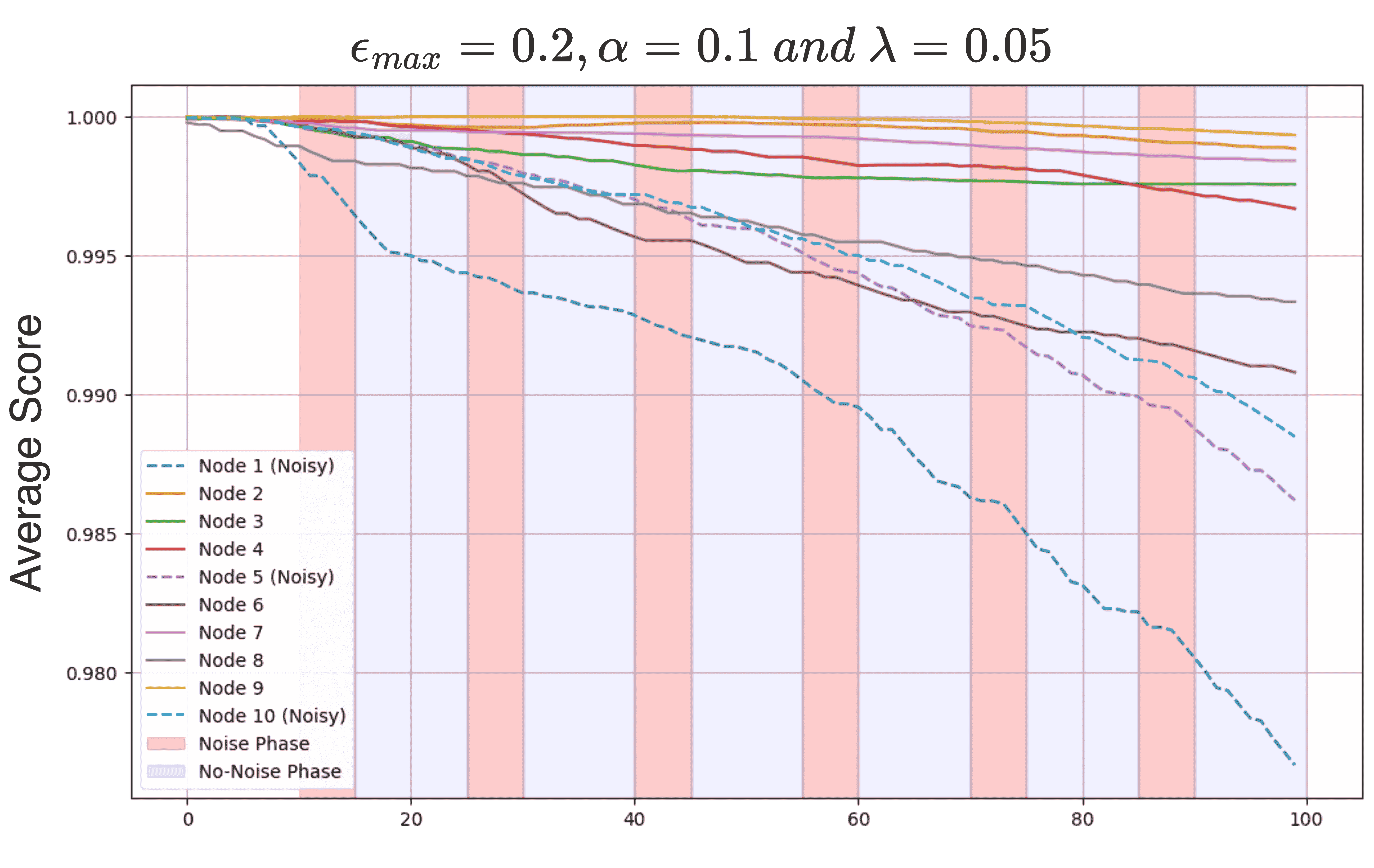}
         &
         \includegraphics[width=0.48\linewidth]{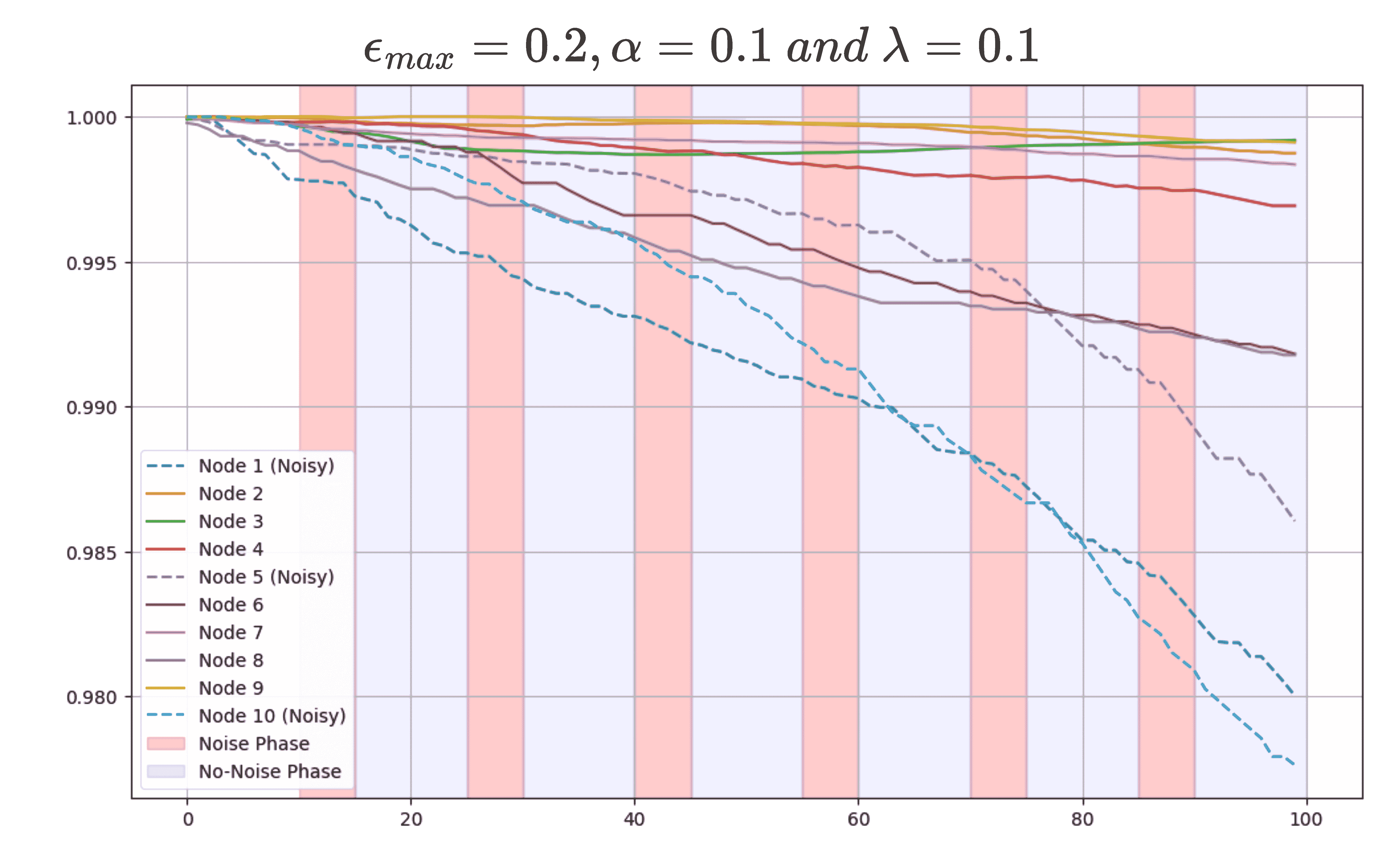}
         \\
         \includegraphics[width=0.48\linewidth]{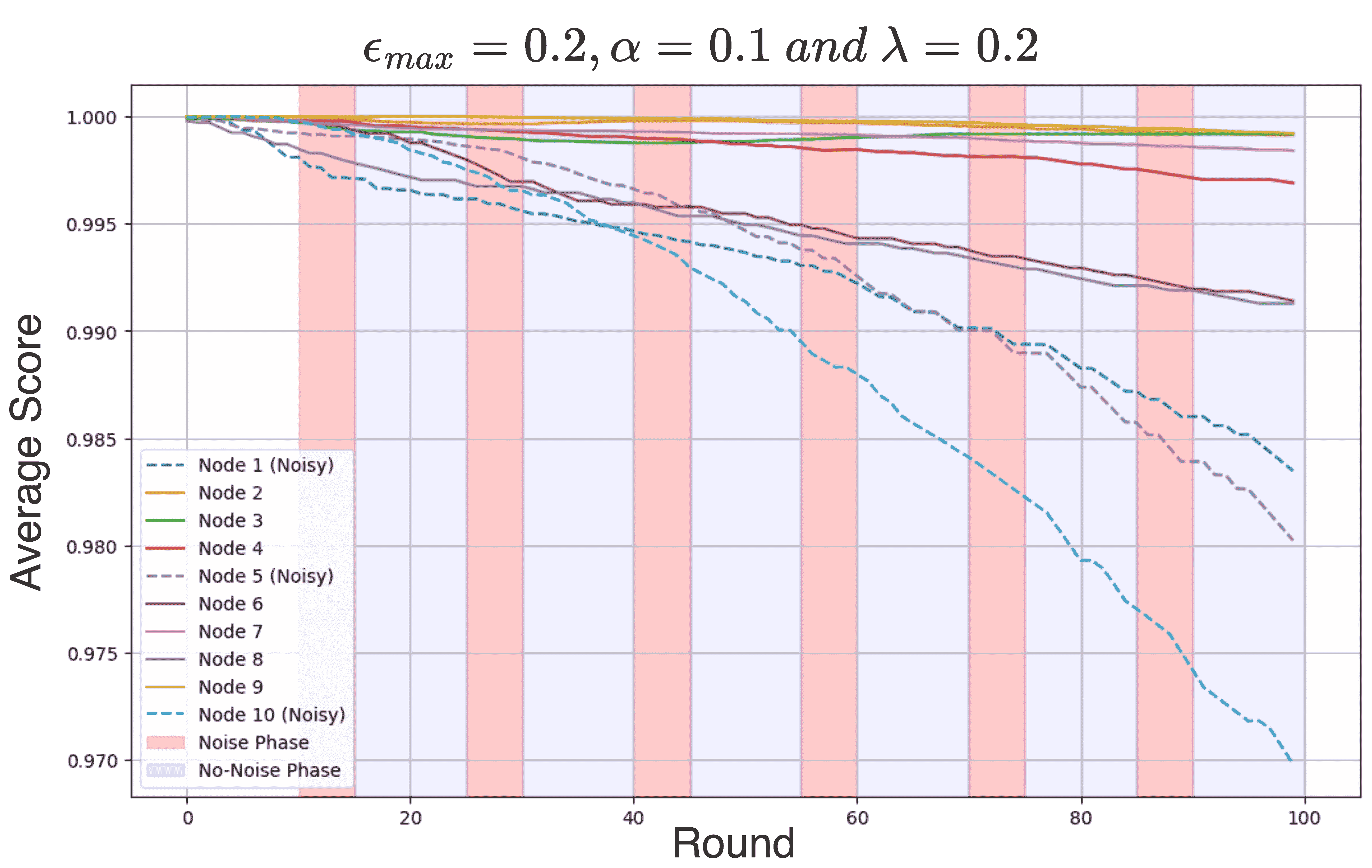}
         &
         \includegraphics[width=0.48\linewidth]{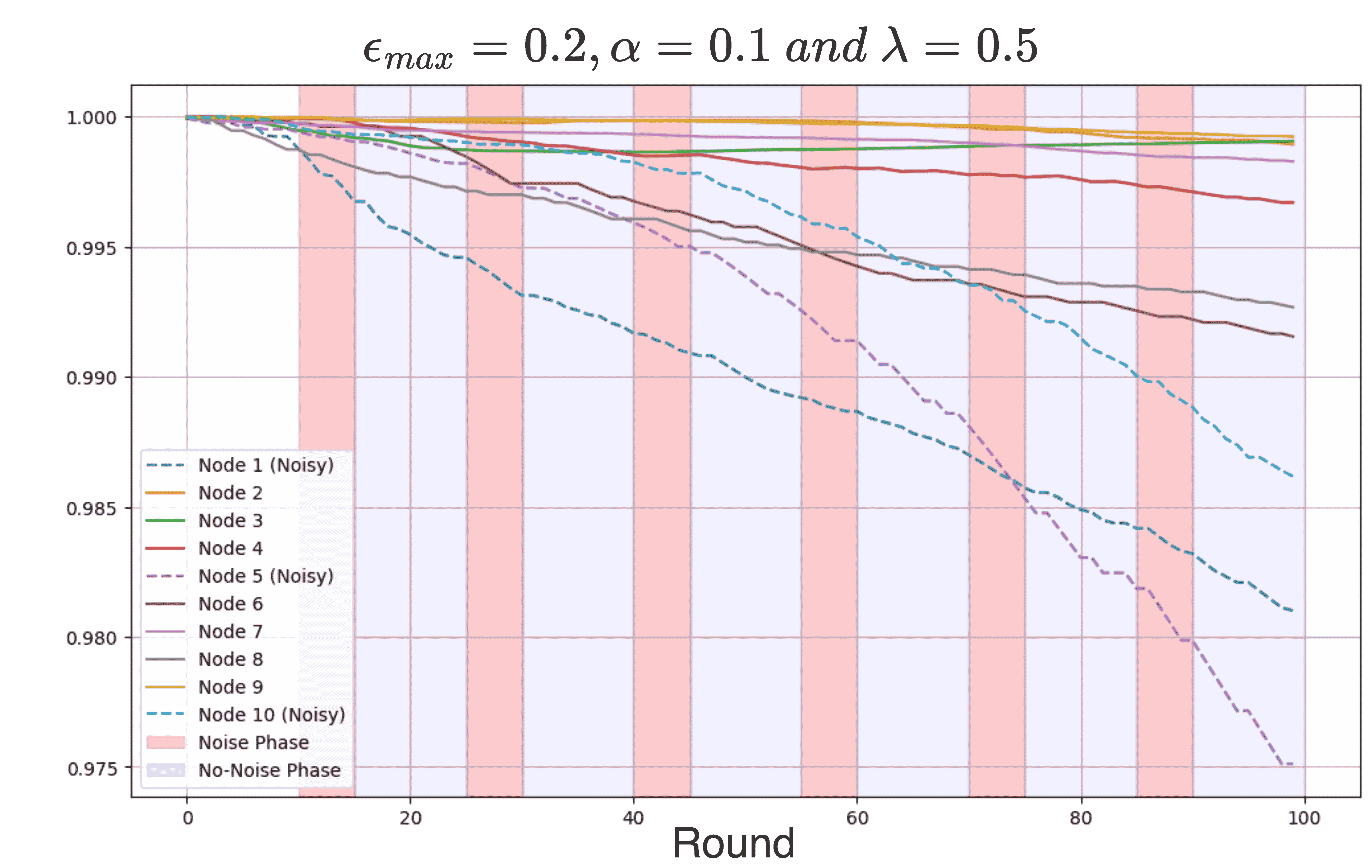}
    \end{tabular}

    \caption{Reputation score evolution over rounds for different values of the importance of intra-cluster concordance index on clustering (\(\lambda\)). The four plots correspond to \(\lambda = 0.05\), \(\lambda = 0.1\), \(\lambda = 0.2\), and \(\lambda = 0.5\), respectively.}
    \label{fig:lambda_reputation}
\end{figure*}

\subsubsection{\textbf{Impact of $\lambda$ on reputation Score Evolution}}
The results in Figure~\ref{fig:lambda_reputation} depict how different values of \(\lambda\) (which, as stated in~\ref{sec:form}, controls the importance of intra-cluster concordance index) affect reputation score evolution over multiple rounds.
\noindent At a low \(\lambda\) value (\(0.05\)), the reputation scores of both normal and noisy nodes degrade gradually. The system remains tolerant to reputation fluctuations, allowing noisy nodes to maintain their position for a longer period. While this helps to avoid false penalties on regular nodes, it also delays the detection and suppression of noisy nodes, potentially leading to prolonged misinformation spread. For \(\lambda = 0.1\), we observe a balance between stability and adaptability.
Noisy nodes gradually lose reputation over time, while regular nodes remain stable. This setting is useful in distinguishing between reliable and unreliable nodes without introducing excessive volatility in the reputation scores. Lastly, higher \(\lambda\) values (\(0.2 \) and \(0.5\)) enforce a more aggressive clustering policy. Noisy nodes experience a relatively more rapid decline in reputation. This strict filtering process may also increase the risk of over-penalizing legitimate nodes, particularly in cases where reputation fluctuations are temporary. The selection of \(\lambda\) is crucial in determining the trade-off between stability and responsiveness. While lower values of \(\lambda\) ensure a smoother reputation evolution process, higher values enhance noise suppression but risk false detections. An optimal choice of \(\lambda\) should consider the required level of noise tolerance, balancing adaptability and robustness in the identification of noisy clients and concordance in survival prediction.

\subsubsection{\textbf{Effect of the Duration of Honest Reporting on Model Accuracy and Reputation Stability}}

Table~\ref{tab:honesty} summarizes the impact of varying the honest period \(T_{\text{honest}}\) on global model accuracy and reputation stability. As \(T_{\text{honest}}\) increases, the global model accuracy improves, while the reputation score stability (measured as the variance of reputation scores across nodes) decreases, indicating that the discriminative power of the model increases as the injected noise is more abrupt.

\begin{table}[ht]
    \centering
       \caption{Impact of Honest Period \(T_{\text{honest}}\) on Global Model Accuracy and reputation Stability.}
    \label{tab:honesty}
    \begin{tabular}{lcc}
     
        \toprule
        \(T_{\text{honest}}\) value & Global C\_Index & reputation Stability \\
        \midrule
        5  & 0.6030 & 0.0006 \\
        10 & 0.6215 & 0.0004 \\
        20 & 0.6351 & 0.0003 \\
        \bottomrule
    \end{tabular}
\end{table}

\subsubsection{\textbf{Impact of reputation Update Frequency}}
\label{subsec:reputation_freq_experiment}

\begin{figure}[ht]
    \centering
    \includegraphics[width=0.48\textwidth]{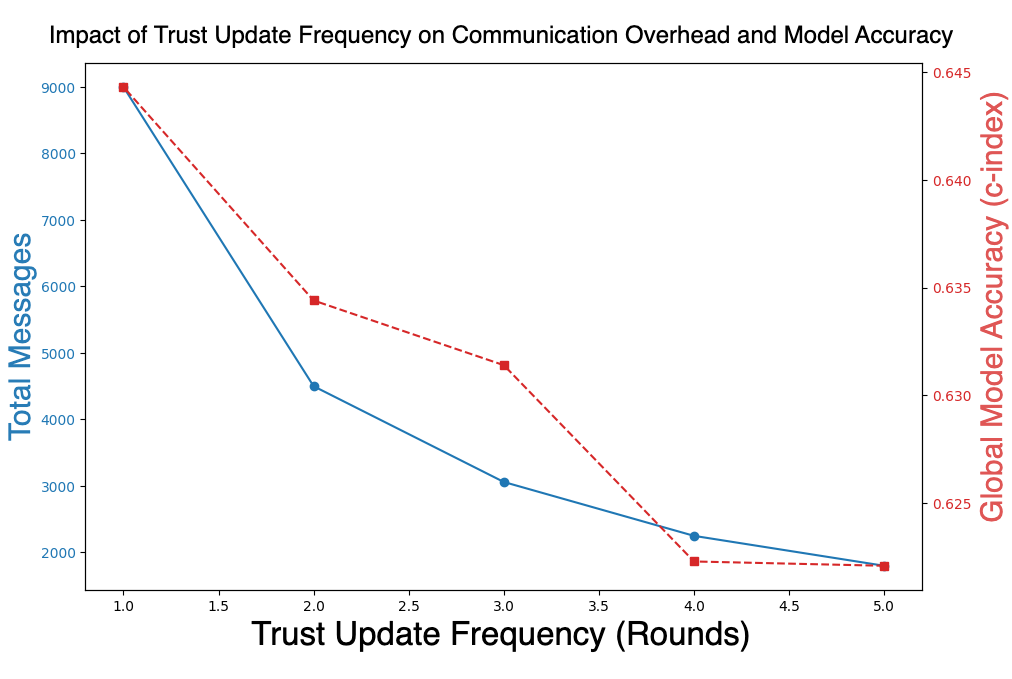}
    \caption{Impact of varying reputation update frequency (\(f\)) on communication overhead (left \(y\)-axis) and global model accuracy (right \(y\)-axis).}
    \label{fig:freq_experiment} 
\end{figure}

In this experiment, we investigate how varying the frequency of reputation updates influences both communication overhead and model performance in our federated learning setup. Specifically, we vary the reputation update frequency (\(f\)) across low, medium, and high regimes, and measure two key metrics: 1) \textit{Message Overhead.} The total number of reputation-related messages exchanged, and 2) \textit{Global Model Accuracy}. Evaluated using the concordance index (C-index).
Figure~\ref{fig:freq_experiment} presents a dual-axis plot of these metrics, with the left \(y\)-axis denoting the total messages and the right \(y\)-axis the global model accuracy. The message passing overhead at the client level is given by \textit{(number of clients in the same cluster - 1)} \( \times \) \( \lfloor \) \textit{total time/frequency of updates} \( \rfloor \), which we have now explicitly added to the paper. As the reputation update frequency increases, the communication overhead (blue curve) drops significantly, reducing the total number of messages. However, this reduction in overhead coincides with a moderate decline in the global model accuracy (red curve). Consequently, there is a trade-off between communication efficiency and model performance: while fewer reputation updates lower overhead, overly infrequent updates may cause the federated model to rely on outdated or less precise scores. Balancing these competing objectives involves choosing an intermediate update frequency that provides accuracy at a reasonable communication cost.

\subsection{Static Noise Bias Injection}
\label{subsec:static_noise_bias_experiment}

\begin{figure}[ht]
    \centering
    
    \includegraphics[width=0.48\textwidth]{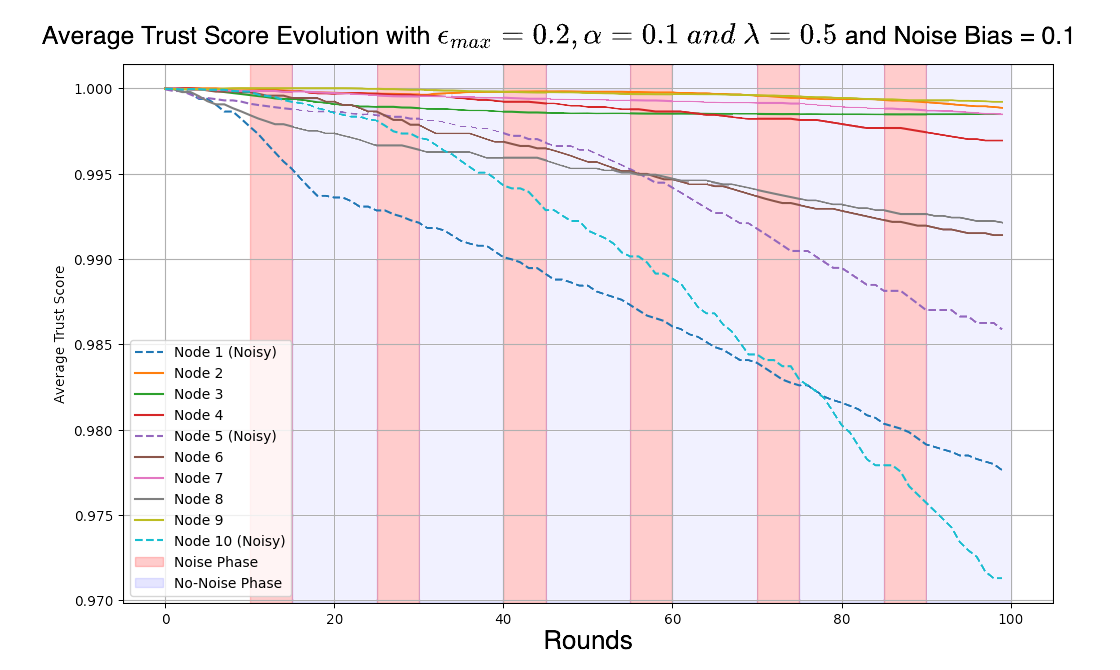}
    
    \caption{Average reputation score evolution with a static noise bias of \(0.1\). Despite the constant shift in malicious updates, the framework effectively detects and isolates dishonest nodes (dashed lines).}
    
    \label{fig:static_noise_bias}
\end{figure}

In this experiment, each malicious node injects noise drawn from a distribution with a constant bias of \(0.1\). Unlike zero-mean noise, this biased noise consistently shifts the node's updates, making adversarial behavior difficult to detect. Figure~\ref{fig:static_noise_bias} illustrates the evolution of the average reputation score for each node over 100 rounds, with pink-shaded regions indicating noise phases. Despite the persistent bias, the proposed reputation mechanism still identifies and penalizes dishonest nodes (dashed lines), gradually reducing their reputation scores while preserving a high reputation for honest nodes (solid lines). These results highlight the resilience of our approach, which dynamically down-weights adversarial contributions even under static, non-zero mean noise conditions.

\subsubsection{\textbf{Robustness Evaluation: Our Model vs. TFFL \& No-Reputation Aggregation}}
\label{subsubsec:robustness_evaluation}

To evaluate the robustness of our reputation mechanism, we conducted experiments in a survival analysis setting where each client trains a local Cox Proportional Hazards (CoxPH) model (see Section~\ref{sec:coxph}) using an increasing fraction of available data over 10 rounds. We compared our proposed peer-driven reputation model with two other aggregation strategies:

\begin{itemize}
    \item \textbf{TFFL Baseline:} This baseline implements the Trustworthy and Fair Federated Learning (TFFL) framework proposed by Rashid et al.~\cite{rashid2025trustworthy}. TFFL employs a Dynamic Reputation-based Consensus (DRC) mechanism that computes client reputation scores using Hybrid Subjective Logic (HSL). Aggregation uses Federated Averaging (FedAvg), weighted by reputation scores.
    \item \textbf{No Reputation:} Client updates are aggregated using standard Federated Averaging (FedAvg)~\cite{mcmahan2017communication} without reputation-based weighting.
\end{itemize}

The global model is evaluated on a held-out global evaluation dataset, and the global concordance index (C-index) is computed for each round. Table~\ref{tab:global_performance} summarizes the global C-index values over 10 rounds. Our model achieves a high and stable C-index, ranging from 0.6573 to 0.6612, consistently outperforming the no-reputation method. The TFFL baseline shows competitive performance, occasionally achieving higher values (e.g., 0.6701 in round 7), but its reliance on server-driven consensus and blockchain overhead leads to greater variability. Our model’s peer-driven reputation mechanism, combined with clustering (Section~\ref{sec:form}), effectively detects and down-weighs noisy clients, ensuring high-quality updates dominate the global model aggregation.

\begin{table}[ht]
\caption{Global C-index over 10 rounds for the three aggregation strategies.}
\label{tab:global_performance}
\centering
\begin{tabular}{rrrr}
\toprule
Round & Our\_Model & TFFL\_Baseline & No\_reputation \\
\midrule
1  & \textbf{0.661218} & 0.6121 & 0.660216 \\
2  & \textbf{0.659425} & 0.605211 & 0.641424 \\
3  & \textbf{0.657350} & 0.624434 & 0.637348 \\
4  & \textbf{0.658202} & 0.621327 & 0.647202 \\
5  & \textbf{0.658222} & 0.610206 & 0.628222 \\
6  & \textbf{0.658495} & 0.620947 & 0.638495 \\
7  & 0.658457 & \textbf{0.6701} & 0.655426 \\
8  & \textbf{0.657647} & 0.611321 & 0.637647 \\
9  & \textbf{0.657726} & 0.619019 & 0.621726 \\
10 & \textbf{0.657853} & 0.615897 & 0.647853 \\
\bottomrule
\end{tabular}
\end{table}

\subsection{Real world application}

We evaluated our approach on the SEER dataset~\cite{SEER}, which comprises breast cancer cases from ten states (2000 - 2017). Treating each state as a federated node enabled us to capture data heterogeneity across regions. 
\begin{table}[ht]
\centering
\caption{C-Index comparison across different calculation methods in SEER dataset.}
\label{table:seer_cindex_comparison}
\begin{tabular}{lrrr}
\toprule
\textbf{Dataset} & \textbf{Our\_Model} & \textbf{TFFL} & \textbf{No\_Reputation} \\
\midrule
KY & \textbf{0.73813} & 0.72010 & 0.70591 \\
LA & \textbf{0.73594} & 0.71982 & 0.72768 \\
CT & \textbf{0.73720} & 0.71500 & 0.73390 \\
UT & \textbf{0.72433} & 0.71850 & 0.72369 \\
GA & \textbf{0.72432} & 0.72145 & 0.72038 \\
HI & 0.76109 & \textbf{0.77025} & 0.73105 \\
NJ & \textbf{0.72831} & 0.72030 & 0.71914 \\
NM & \textbf{0.72740} & 0.72310 & 0.71592 \\
IA & \textbf{0.81941} & 0.79040 & 0.79920 \\
CA & \textbf{0.73437} & 0.72560 & 0.72499 \\
\bottomrule
\end{tabular}
\end{table}
As shown in Table~\ref{table:seer_cindex_comparison}, our model generally outperformed the no-reputation baseline and achieved competitive performance relative to the TFFL framework~\cite{rashid2025trustworthy}. For instance, in LA, CT, and UT, our model achieved C-index values of 0.73594, 0.73720, and 0.72433, respectively—values that surpass the corresponding results of the no-reputation method. Although TFFL attained slightly higher C-index values in one state (0.77025 in HI), our approach consistently maintained robust performance across most regions. These results underscore that integrating a peer-driven reputation mechanism with clustering-based noise handling not only improves accuracy but also leverages the diversity inherent in the federated SEER data.

\section{Conclusion}

In this paper, we propose a robust, peer-driven reputation mechanism for federated learning (FL) in healthcare. We introduce a customized threat model that captures nodes gradually injecting noise to evade detection, leveraging differential privacy to decouple reputation computation from federated aggregation: privatized client updates are used exclusively for peer evaluation, while unaltered updates inform global model training. Our experimental results, supported by simulation studies and a real-world evaluation of the SEER dataset, demonstrate that integrating dynamic peer feedback with clustering-based noise handling enhances the robustness and accuracy of survival prediction models, by down-weighing noisy or misbehaving client updates, as evidenced by the evolution of reputation scores under varying noise injection scenarios, and leveraged the inherent heterogeneity across federated nodes to improve generalizability while preventing the breach of sensitive healthcare data during model aggregation.

Our framework consistently achieved high concordance index values over multiple training rounds, outperforming the no-reputation baseline while exhibiting competitive performance relative to the Trustworthy and Fair Federated Learning framework, which employs a dynamic reputation-based consensus mechanism with hybrid subjective logic to compute reputation scores, adapting to client behavior through temporal dynamics and multi-factor evaluations. As part of our ongoing research, we are investigating how FL aggregation is affected by different clinical datasets characterized by varying data distributions and how message overheads for reputation updates scale in larger networks with more clients. Furthermore, we plan to explore novel techniques to analyze and quantify the impact of feature space variations on patient outcomes, providing valuable insights to hospitals by identifying influential clinical features and guiding improvements in patient care.

\section*{References}
\vspace{-6mm}
\bibliographystyle{IEEEtran}
\bibliography{main}

\end{document}